\documentclass[pdflatex,sn-mathphys-num]{sn-jnl}

\usepackage{graphicx}
\usepackage{multirow}
\usepackage{amsmath,amssymb}
\usepackage{textcomp}
\usepackage{booktabs}
\usepackage{siunitx}
\usepackage[section]{placeins}
\usepackage{tikz}
\usetikzlibrary{calc,arrows.meta,positioning}
\usepackage{bm}
\newcounter{frameno}
\newcommand{\labelframe}[3]{%
  \begin{tikzpicture}[inner sep=0,outer sep=0]
    \node[anchor=south west] (im) {\includegraphics[#2]{#3}};
    \node[anchor=north west,fill=black,text=white,inner sep=1.5pt,
          font=\sffamily\footnotesize\bfseries]
      at ([xshift=1.5pt,yshift=-1.5pt]im.north west) {#1};
  \end{tikzpicture}}

\begin{document}

\title[Long-Distance Real-World Navigation]{Long-Distance Real-World Navigation of the Legged-Wheeled Robot Go2-W Using Deep Reinforcement Learning}

\author[1]{\fnm{Takaaki} \sur{Matsuzawa}}
\author*[1]{\fnm{Kiyoshi} \sur{Irie}}\email{irie@furo.org}
\author[1]{\fnm{Tomoaki} \sur{Yoshida}}
\author[1]{\fnm{Taro} \sur{Suzuki}}
\author[1]{\fnm{Yoshitaka} \sur{Hara}}
\author[1]{\fnm{Masahiro} \sur{Tomono}}

\affil*[1]{\orgdiv{Future Robotics Technology Center (fuRo)}, \orgname{Chiba Institute of Technology}, \orgaddress{\city{Narashino}, \state{Chiba}, \country{Japan}}}

\abstract{Legged-wheeled robots have been extensively studied for their potential to combine the efficient flat-ground mobility of wheels with the rough-terrain adaptability of legs.
However, their application to long-distance autonomous navigation in real environments remains limited.
This paper reports our efforts to build a deep reinforcement learning-based locomotion controller and an autonomous navigation system for the commercially available Go2-W legged-wheeled robot.  The study focuses on the integrated system and the practical challenges associated with its sustained operation over long distances in real-world environments.
For locomotion control, we extended our proprioception-only policy, previously developed for quadruped robots, to Go2-W, a 16-degree-of-freedom legged-wheeled robot.
We additionally found that wheeled locomotion concentrates the load on the hip joints, resulting in heat concentration that hinders sustained travel.
To address this issue, we developed a policy with periodic foot-lifting motion that distributes the joint loads and mitigates heat concentration.
We evaluated the integrated system at the Tsukuba Challenge 2025. The robot autonomously traversed an approximately \SI{2.8}{\kilo\meter} route that included sidewalks, a park, and stairs without stopping owing to overheating.}

\keywords{legged-wheeled robot, deep reinforcement learning, autonomous navigation, locomotion control, Tsukuba Challenge}

\maketitle

\section{Introduction}

Legged-wheeled robots have been studied as a promising configuration that can potentially combine efficient wheeled locomotion on flat terrain with the ability to traverse rough terrain using legs~\cite{kiybib:bjelonic_clawar_2022_survey}.
However, the mechanical and control complexities arising from the simultaneous use of legs and wheels remain challenging.

Various configurations of legged-wheeled robots have been proposed~\cite{jkiybib:kimura_jrsj_1992,kiybib:endo_advanced_robotics_2012_roller_walker},
and recent advances in actuators, computing, and control technology have enabled real-world applications of legged-wheeled robots~\cite{kiybib:bjelonic_iros_2021_mpc,kiybib:chamorro_icra2024_stair}.
Commercially available robots have also emerged~\cite{wl:unitree_go2w,wl:ascento_guard}, creating an environment in which researchers and developers can readily employ legged-wheeled platforms.
Nevertheless, most of the literature on legged-wheeled robots focuses on mechanism design and locomotion control, whereas reports on their deployment in practical mobile-robot tasks remain limited.

To our knowledge, the only published research demonstrating long-distance outdoor navigation with a legged-wheeled robot is that of Lee et al.~\cite{kiybib:lee_science_robotics_2024}; however, instances of human intervention during the operation were reported. Thus, a fully autonomous long-duration, long-distance travel remains an open challenge.

We have previously researched deep reinforcement learning (DRL)-based locomotion control and autonomous navigation for quadruped robots~\cite{kiybib:irie_ar_2025_qrc}.
In this study, we extend this prior research to the commercially available Go2-W legged-wheeled robot and report the results of long-distance autonomous navigation in real-world environments.

This paper does not highlight the novelty of individual components. Instead, it focuses on the practical problems that arise when a commercially available robot is operated for long durations and over a long distance in a real environment, and on how they can be addressed at the overall system level.

The main contributions of this paper are as follows.
\begin{enumerate}
  \item We present an integrated outdoor autonomous navigation system built around a commercially available legged-wheeled robot and demonstrate its effectiveness through long-distance autonomous travel in a real-world environment. We developed and used our own DRL-based methods for locomotion and path-following control, and combined them with conventional methods for localization and high-level path planning.
  \item We extend a DRL-based locomotion controller originally developed for quadruped robots to a 16-DoF legged-wheeled robot. We describe the addition of wheel joints to the action space, configurations of the two locomotion policies for flat and rough terrains, and differences in their training settings, and evaluate their performance through simulations and application to a real robot.
  \item We identify a problem in which heat concentrates at the hip joints during wheeled locomotion of the Go2-W. Using temperature-rise and joint-torque measurements, we analyze the causes for the heat concentration and demonstrate that periodic foot-lifting motion suppresses overheating, thereby enabling long-duration locomotion.
\end{enumerate}

The remainder of this paper is organized as follows.
Section~\ref{sec:related_work} reviews the related studies. Section~\ref{sec:system} describes the overall software and hardware architectures of the target system.
Section~\ref{sec:locomotion} demonstrates low-level locomotion control based on DRL, and Section~\ref{sec:navigation} describes the higher-level software system that realizes autonomous navigation.
Section~\ref{sec:experiment} presents the experimental results obtained in real-world environments. Section~\ref{sec:discussion} discusses the effectiveness of the foot-lifting gait against overheating and the unresolved challenges.
Finally, Section~\ref{sec:conclusion} concludes the paper.

\section{Related Work}\label{sec:related_work}

\subsection{Development of DRL for Legged Robots}
DRL-based locomotion control for legged robots has advanced rapidly in recent years.
This study builds upon these developments.
A primary challenge in applying DRL to the motion control of actual robots has been the large number of trials required for training, which renders direct on-robot learning impractical.
To address this, sim-to-real frameworks that transfer policies trained in simulation to real robots have been explored through various approaches~\cite{Tan-RSS-18,Hwangbo2019science}.

These sim-to-real methods have further been extended to traversal through diverse rough terrains.
Lee et al. achieved terrain-adaptive locomotion with a blind (proprioceptive) policy by distilling the knowledge of a teacher policy trained with privileged information into a student policy that operates solely on proprioceptive sensing~\cite{kiybib:lee_science2020}.
Kumar et al. proposed a method for rapid online adaptation to changing dynamics, including varying terrain properties~\cite{Kumar2021RMA}.
Miki et al. combined this approach with exteroceptive sensing, such as terrain elevation maps, and demonstrated robust locomotion control in natural environments~\cite{kiybib:Miki2022}.

The computational infrastructure supporting such learning has also been advancing.
Recent advances in GPU-accelerated simulation frameworks have enabled parallel training across many environments, substantially shortening the policy learning process.
Representative examples include Isaac Gym~\cite{IsaacGym}, a GPU-based parallel physics simulator, and Legged Gym~\cite{RudinHR021}, which builds on this capability to train locomotion policies for legged robots. More recently, MuJoCo Playground~\cite{mujoco_playground}, which extends the MuJoCo physics engine with GPU support, has also been released.
These frameworks have been widely adopted, and research addressing challenging tasks such as acrobatic maneuvers and locomotion in narrow spaces is progressing~\cite{xu_icra2024_confined,luo_ral2024_pie,xiao_npj2025_bipedal}.

Through these developments, DRL has emerged as an effective approach for the locomotion control of legged robots.
We have also been conducting research on proprioceptive policies for commercially available quadruped robots~\cite{kiybib:irie_ar_2025_qrc}.
In this study, we extend these results to a legged-wheeled robot.

\subsection{Configurations of Legged-Wheeled Robots}
Legged robots possess high rough-terrain traversability, whereas wheeled robots have superior locomotion speed and energy efficiency on flat surfaces.
To combine the advantages of both locomotion modes, legged-wheeled robots equipped with wheels at the ends of their legs have been studied for decades~\cite{kiybib:bjelonic_clawar_2022_survey}.
Various configurations have been proposed for legged-wheeled robots, which differ in the number of legs, role of wheels, and other design aspects.

A representative bipedal configuration comprises two legs with driven wheels at the feet, where the wheels are used for balancing while moving, such as Ascento by Klemm et al.~\cite{wl:klemm_icra2019_ascento,wl:klemm_ral_2020_ascento_lqr}.
Quadruped configurations with four wheels are statically stable and can leverage the redundant legs for superior step-traversal capability.
Examples include the ANYmal-based robot with wheels, developed by Bjelonic et al.~\cite{kiybib:bjelonic_ral_2019}, and Go2-W used in this study.
Other configurations include the Roller-Walker, which uses passive wheels and is propelled primarily by leg motions~\cite{kiybib:endo_advanced_robotics_2012_roller_walker};
ASTERISK H, a hexapod equipped with driven wheels~\cite{wl:takubo_jsmec2009_leg_wheel_hybrid};
and configurations combining driven and passive wheels~\cite{wl:oda_jrsj2022_hydraulic_rover}.

\subsection{Control and Applications of Four-Wheeled Legged-Wheeled Robots}
Legged-wheeled robots must simultaneously handle discrete contact transitions  of the legs and continuous rolling motion by the wheels.
This leads to control problems that differ from those of legged-only or wheeled-only robots.
In classical studies, wheel locomotion control and leg balance control were designed separately~\cite{wl:besseron_iros2008_hylos}.
Subsequently, whole-body motion control accounting for the kinematic constraints between legs and wheels was proposed, enabling integrated locomotion combining wheeled driving and leg motion~\cite{kiybib:bjelonic_ral_2019}.
Furthermore, model predictive control (MPC)-based methods that predict and optimize whole-body motion and contact forces have been proposed, realizing hybrid locomotion that combines wheeled driving and walking depending on the situation~\cite{kiybib:bjelonic_iros_2021_mpc}.
However, the computational cost of MPC tends to scale cubically with the state and input dimensions. Reducing this burden remains an open issue.

In model-based control, contact state transitions and interactions with the terrain must be explicitly modeled, and improving the adaptability of the robot to rough terrain has been challenging.
Recently, as with the legged robots discussed in the previous subsection, DRL has been applied to the locomotion control of legged-wheeled robots.
Chamorro et al. used reinforcement learning to acquire a proprioceptive policy for legged and legged-wheeled robots. The approach was demonstrated on different types of robots through simulation and on Ascento in the real world~\cite{kiybib:chamorro_icra2024_stair}, demonstrating that rough-terrain traversal is possible even with a proprioceptive policy.
DRL requires only the execution of a learned policy function at runtime; hence, the computational cost on a real robot is relatively low.

Lee et al. further proposed a learning-based locomotion control approach that incorporates terrain perception from exteroceptive sensors as additional policy inputs.
By integrating this with a navigation system, they conducted autonomous locomotion experiments covering a cumulative distance of \SI{8.3}{\kilo\meter}, including rough terrain, across an area spanning $245\times345$\,\si{\meter} in an urban environment~\cite{kiybib:lee_science_robotics_2024}.
This study demonstrated the feasibility of long-distance autonomous locomotion using a legged-wheeled robot. It also identified three types of situations that required human intervention: precautionary safety stops, path-planning failures caused by environmental changes, and localization errors.

\section{Robot System}\label{sec:system}

This section describes the hardware and software configurations of the developed autonomous mobile system.
The proposed system is an extension of the autonomous mobile system based on the quadruped robot Go2 used in our prior study~\cite{kiybib:irie_ar_2025_qrc}.
Although many components are common between the original and extended systems, the entire system is described here for completeness.

\subsection{Hardware Configuration}

\begin{figure}[tb]
  \centering
  \includegraphics[width=0.75\linewidth]{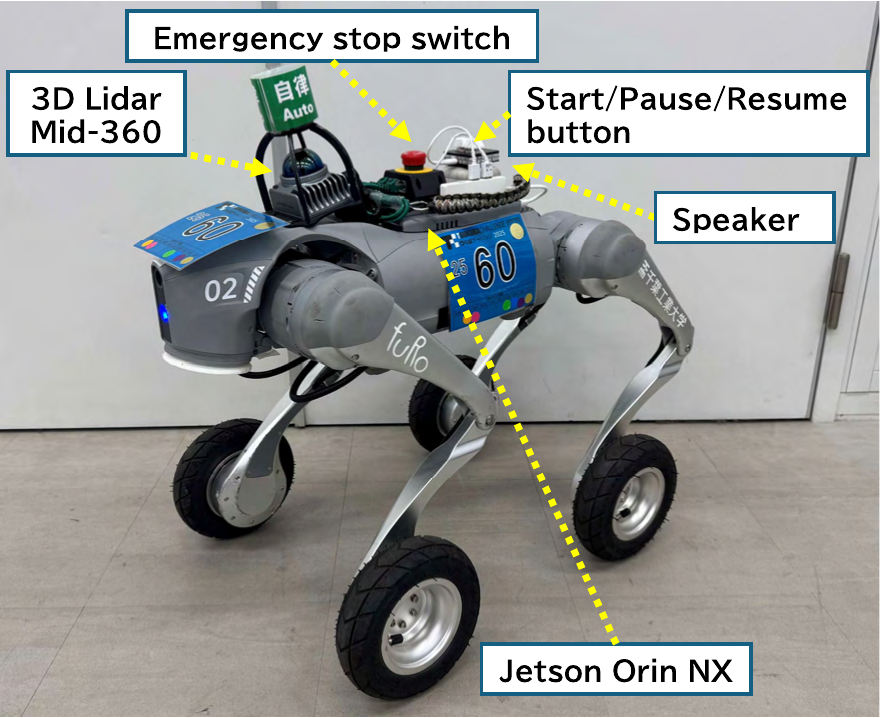}
  \caption{Overview of our robot system based on the Unitree Go2-W. The system is equipped with a Livox MID-360 3D Lidar, an NVIDIA Jetson Orin NX, an emergency stop button, and a USB speaker.}
  \label{fig:go2w_outlook}
\end{figure}

\begin{figure}[tb]
  \centering
  \includegraphics[width=0.75\linewidth]{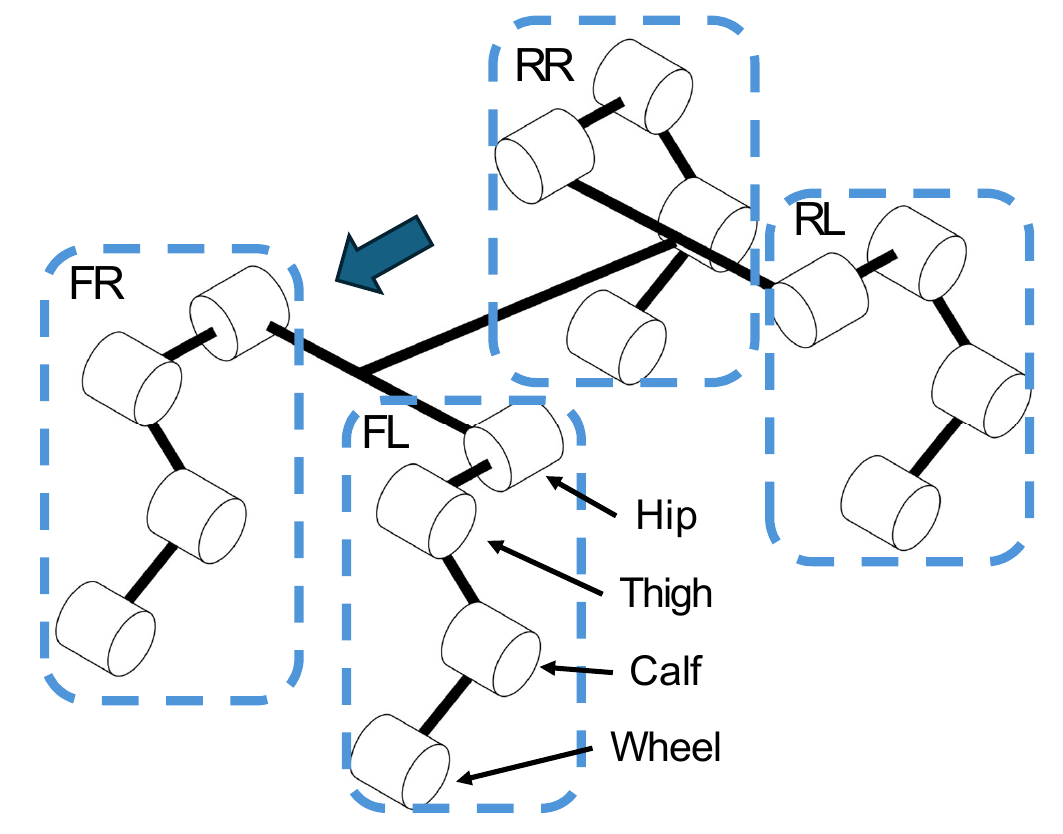}
  \caption{Degrees of freedom and joint names of the Go2-W. Each leg has four joints: hip, thigh, calf, and wheel, totaling 16 DoF.}
  \label{fig:go2w_dof}
\end{figure}

This study used Unitree Go2-W, a legged-wheeled robot, as the base platform.
Go2-W is derived from the quadruped robot Go2, which has 12 leg-joint axes, by replacing the feet with wheels, resulting in a 16-degree-of-freedom legged-wheeled robot with four additional wheel axes.
The appearance of the developed robot system is shown in Fig.~\ref{fig:go2w_outlook}; Fig.~\ref{fig:go2w_dof} shows the degrees of freedom and naming convention followed for the joint; and Table~\ref{tab:go2w_hardware} lists the system specifications.
The legs have no rotational degree of freedom about the vertical axis, preventing the steering of the wheels.
Therefore, the robot lifts and repositions its feet to move laterally or to turn.

\begin{table}[tb]
  \centering
  \caption{Specifications of our robot system}
  \label{tab:go2w_hardware}
  \small
  \begin{tabular}{ll}
    \toprule
    Base robot & Unitree Go2-W (R\&D edition) \\
    Size (standing) & \qtyproduct{0.7 x 0.5 x 0.7}{m} \\
    Weight & \qty{22}{kg} \\
    Battery & \qty{29.6}{V}, \qty{15}{Ah} \\
    Joints & 16 DoF (12 leg joints, 4 wheels) \\
    Wheel diameter & \qty{18}{cm} \\
    Main computer & NVIDIA Jetson Orin NX \\
    External sensor & Livox MID-360 3D Lidar \\
    \botrule
  \end{tabular}
\end{table}

A Livox MID-360 3D Lidar is used as the external sensor.
Go2-W is equipped with a rotating 3D Lidar (L1) on its head unit. However, we removed it, as its exposed spinning parts pose a safety risk during experiments on public roads.
The computations required for locomotion control and navigation are performed on an NVIDIA Jetson Orin NX mounted in the rear expansion dock.
Livox MID-360 is used for self-localization, map matching, and obstacle detection.
Additionally, a USB speaker for audible state notifications during debugging and an emergency stop button are mounted on the robot.
The emergency stop button was installed in accordance with safety regulations for experiments on public roads; when pressed, the control software reads its state and halts the robot by setting the velocity command to zero.

\subsection{Software Overview}

\begin{figure}[tb]
  \centering
  \includegraphics[width=0.75\linewidth]{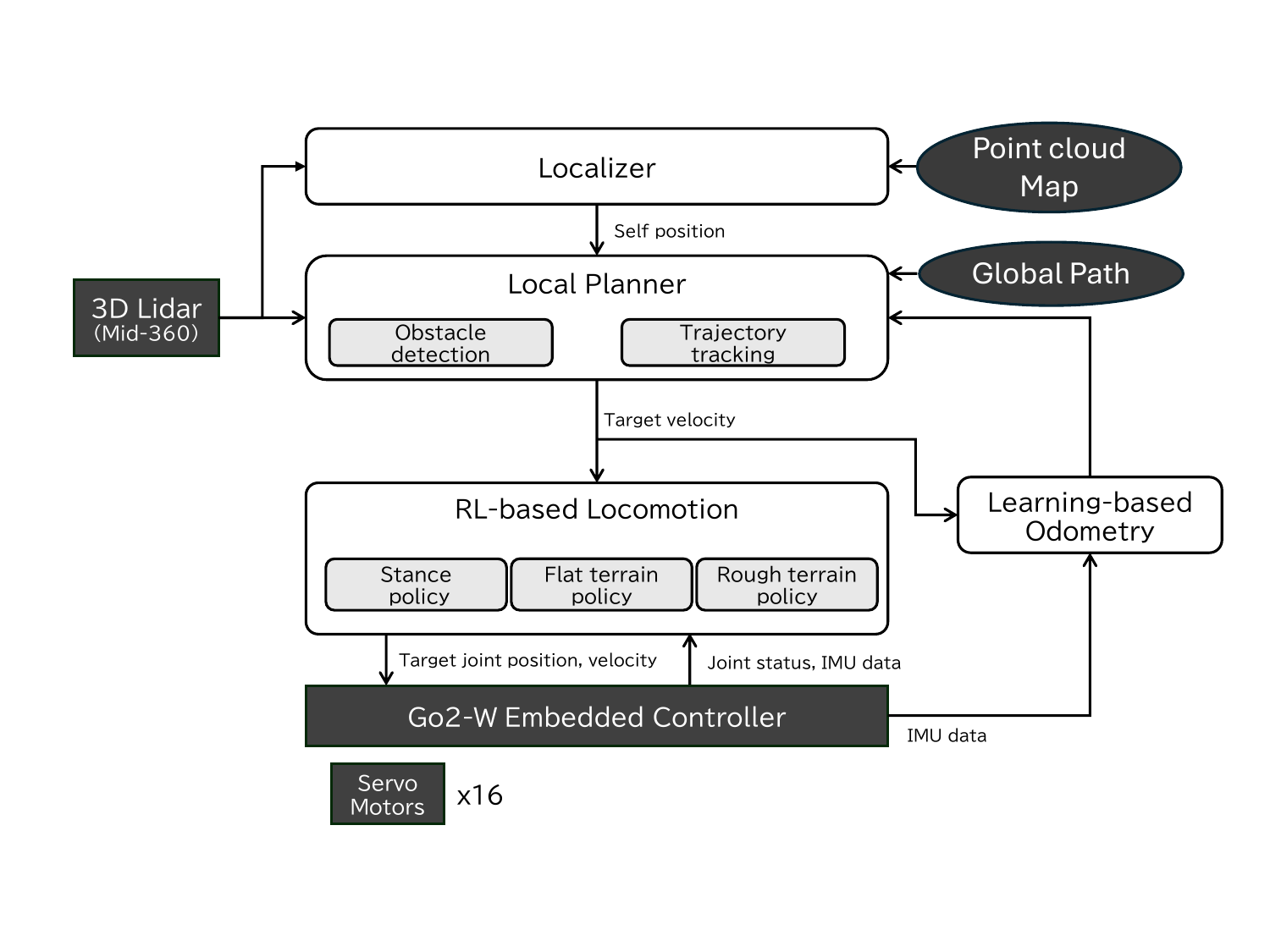}
  \caption{Software architecture of the navigation system.}
  \label{fig:system_architecture}
\end{figure}

Fig.~\ref{fig:system_architecture} shows the main functional architecture of the software.
The system comprises an upper navigation layer and a lower locomotion control layer that computes the commands for the joints.

The upper layer estimates the robot pose on the map using 3D Lidar point clouds, and performs obstacle detection, avoidance, and path following based on the estimation results, sending a target velocity to the lower layer.
Path following also uses a learning-based odometry that estimates the local movement from the velocity command history and inertial measurement unit (IMU) data.
The upper layer is described in Section~\ref{sec:navigation}.

The locomotion control layer employs policies acquired through DRL.
The policy functions take the internal state of the robot, including the joint status and IMU, together with the velocity commands as the input, and output the target angle for each leg joint and target speed for each wheel.
These commands are transmitted via Ethernet to the embedded controller inside the robot through the Unitree SDK.
Position control is applied to the leg joints, and speed control is applied to the wheels.
The system switches between three policies---stance, flat-terrain locomotion, and rough-terrain locomotion---depending on the situation.
Locomotion control is described in Section~\ref{sec:locomotion}, and policy switching is described in Section~\ref{sec:navigation_strategy}.

The onboard computer runs Ubuntu 20.04, and each module is implemented as an independent process.
Inter-process communication is handled by a lightweight shared-memory-based middleware developed by the authors, and a part of the self-localization runs on ROS Noetic.

\section{Locomotion Control}\label{sec:locomotion}

Although Go2-W is equipped with a built-in locomotion controller provided by the manufacturer, extended continuous operation caused heat concentration in certain motors, eventually triggering thermal shutdown, as described in Section~\ref{sec:exp_heat}.
Therefore, we adopted three policies obtained through DRL for locomotion control: stance, flat-terrain locomotion, and rough-terrain locomotion.
The three types of policies could be switched according to the commands from the higher-level layer.
The following subsections present an overview and the training procedure for each of these policies.

\subsection{Policy and Training Overview}

The locomotion policies were trained using an extended version of Legged Gym~\cite{RudinHR021}.
In the original Legged Gym, the policy input includes quantities that are difficult to estimate on a real robot, such as translational velocity and the surrounding terrain information, rendering it infeasible to directly deploy the trained policy on a real robot.
In this study, we realized a proprioceptive policy that achieves stable locomotion by following the same approach as in~\cite{kiybib:irie_ar_2025_qrc}.

\begin{figure}
  \centering
  \includegraphics[width=0.75\linewidth]{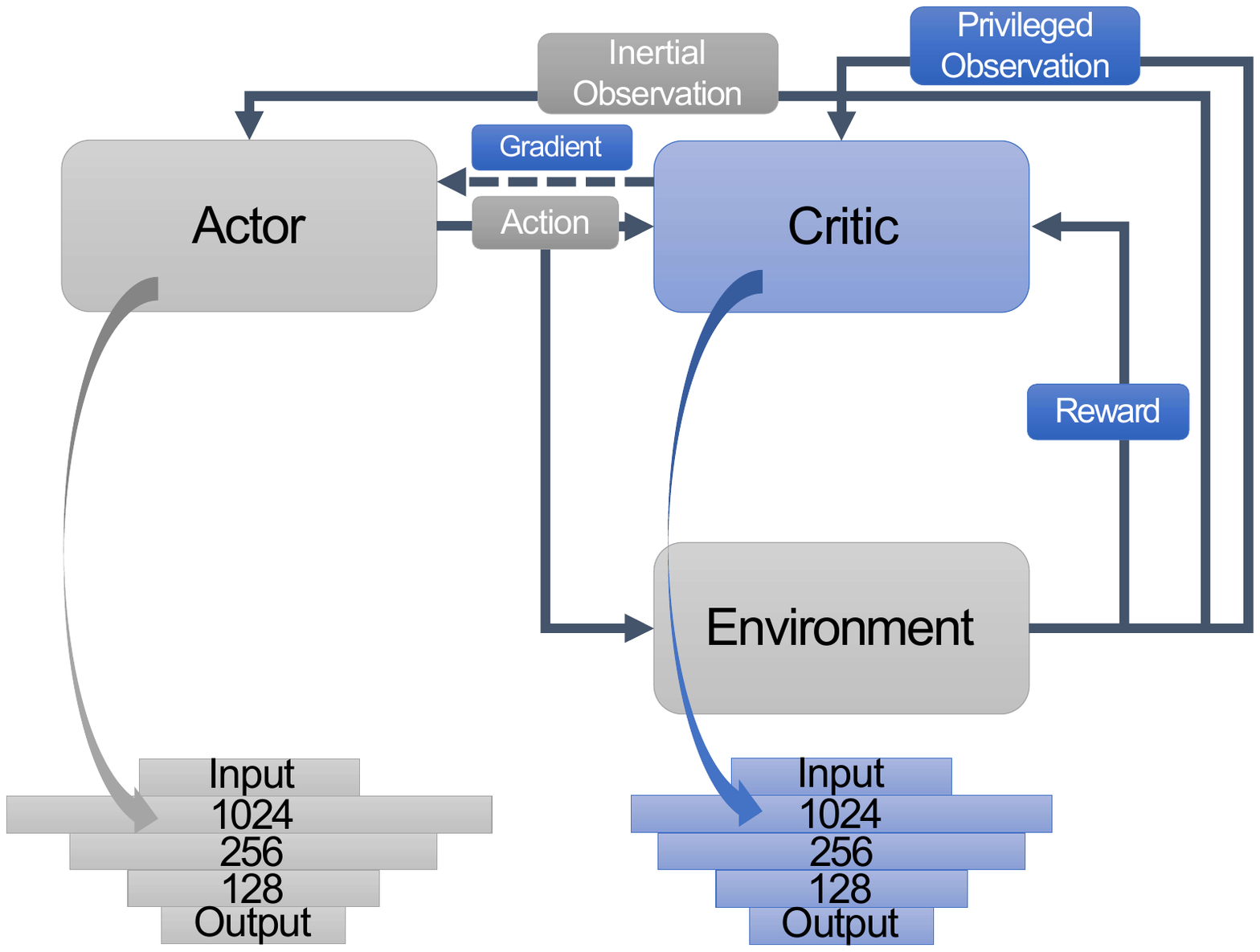}
  \caption{Training framework based on Asymmetric Actor Critic. The critic receives all observations including privileged information available only in simulation, while the actor uses only proprioceptive observations obtainable on the real robot. This enables direct sim-to-real transfer of the trained policy.}
  \label{fig:asymmetric_actorcritic}
\end{figure}

Fig.~\ref{fig:asymmetric_actorcritic} shows the training framework and architectures of the policy and critic functions.
We employed asymmetric actor critic~\cite{Pinto-RSS-18} as the training framework, with proximal policy optimization~\cite{schulman2017proximal} as the learning algorithm.
In this framework, the critic receives privileged observations available only in the simulator, whereas the actor input is restricted to proprioceptive sensor information that can be readily obtained on the real robot.
This enables the policy trained in simulation to be directly deployed on the real robot.
To further bridge the sim-to-real gap, domain randomization~\cite{Tobin2017DomainRF} was employed during the training:
Gaussian noise was added to the sensor observations, and random action delays were introduced.
The training and domain randomization parameters, as well as an ablation study of the reward terms, are presented in Appendix~A.

For all three policy functions described below, a multi-layer perceptron (MLP) with hidden layers of 1024$\times$256$\times$128 and a 16-dimensional output was used with ELU activation functions.
The input consists of the velocity command $(v_x, v_y, \omega)$ from the higher-level layer and robot state observed through various proprioceptive sensors.
The output is the position command (for leg joints) or velocity command (for wheels) to the 16 actuators.

Table~\ref{tab:observations} lists the observations provided as input.
Notably, although the joint positions have 16 dimensions covering both leg joints and wheels, the four wheel-axis joints always receive zero input because they are velocity-controlled.
These 57 dimensions constitute the base observation, to which additional commands and past observations are appended depending on the policy.

\begin{table}[tb]
    \centering
    \caption{Observations used in training control policies.}
    \label{tab:observations}
    \small
    \begin{tabular}{@{}lrc@{}}
        \toprule
        Observation & Dim. & \\
        \midrule
        \multicolumn{3}{l}{\textbf{Proprioceptive}} \\
        \quad Joint positions     & 16 & \\
        \quad Joint velocities    & 16 & \\
        \quad Body angular velocity & 3 & \\
        \quad Projected gravity   & 3  & \\
        \quad Velocity command    & 3  & \\
        \quad Body posture command (for stance policy)   & 3  & \\
        \quad Previous actions    & 16 & \\
        \midrule
        \multicolumn{3}{l}{\textbf{Privileged (training only)}} \\
        \quad Terrain heights      & 374 & \\
        \quad Body linear velocity & 3  & \\
        \quad Friction coefficient & 1  & \\
        \quad Body mass            & 1  & \\
        \quad Action latency       & 1  & \\
        \botrule
    \end{tabular}
\end{table}

\subsection{Stance Policy}

The stance policy was designed to maintain a stable standing posture: all four legs remain in contact with the ground, and the body center is held at a height of \SI{32}{\centi\meter}.
It is used for transitions such as standing up from the lying-down position at startup and stopping from a moving state.

The input to the stance policy has 120 dimensions.
Three posture-command dimensions specifying the desired body pose (roll, pitch, and height) are appended to the 57-dimensional base observation.
One step of past observations is appended to these 60 dimensions, yielding a total input of 120 dimensions.

During the training, the policy learns to transition from various lying-down configurations to the standing posture.
To obtain diverse initial configurations, the robot is dropped from a height (\SI{28}{\centi\meter}) with zero applied torque. The drop produces various lying-down poses with different joint angles without causing the robot to topple over.
Furthermore, to learn safe stopping following locomotion, a separately trained locomotion policy driven by random commands is used to generate mid-locomotion states as the initial configurations.

\subsection{Locomotion Policy} \label{subsec:locomotion_policy}
We prepared two locomotion policies: one for rough terrain and the other for flat terrain.
Both were trained with the primary reward of following the given velocity command $(v_x, v_y, \omega)$.
In addition, as shown in Table~\ref{tab:reward_functions}, the reward function is designed to encourage stability and energy efficiency by employing terms related to the torque magnitude, action variations, and maintenance of a level body orientation.

Legged Gym~\cite{RudinHR021} enables training policies that can handle various terrains within a curriculum learning framework.
Multiple terrain types (flat ground, slopes, stairs, grid-patterned bumps, random bumps, etc.) were generated in the simulation.
At the beginning of the training, the robot was placed on an easy terrain, such as a flat ground; agents that successfully traversed a certain distance were relocated to terrains of greater difficulty or of different types.
Conversely, agents that failed to move ahead (e.g., because of falls) were relocated to easier terrains.
By repeating this process, a locomotion controller adapted to various terrains was learned.

\subsubsection{Rough-Terrain Locomotion Policy}
To handle stairs and steps along the route, we trained a locomotion policy for Go2-W using an environment originally developed for rough-terrain locomotion of quadruped robots~\cite{kiybib:irie_ar_2025_qrc}.
The input comprised 114 dimensions: the 57-dimensional base observation plus one step of past observations.
The resulting policy primarily uses wheels for locomotion. In the evaluation on the real robot, the robot was able to traverse stairs and steps up to \SI{24}{\centi\meter} using this policy, which we considered sufficient for general urban environments.

However, this locomotion policy exhibited load and heat concentration in specific joints, similar to the manufacturer-provided controller, preventing extended operation.
Therefore, we reserved this policy for rough-terrain locomotion such as stairs and trained a separate policy with reduced heating for long-duration operation.

\subsubsection{Flat-Terrain Locomotion Policy}
We observed that wheel-based locomotion caused a load concentration at the hip joints located at the base of each leg, which was the primary cause of heating.
During locomotion, the hip joints gradually spread outward, apparently increasing the load on these joints.
We hypothesized that periodically lifting the feet would reset the hip joints and prevent them from continuing to spread outward.
Building on this hypothesis, we trained a policy that encourages periodic foot-lifting behavior.

The main differences between the training environment used for the rough-terrain locomotion policy and that for the flat-terrain locomotion policy are a reduction in the terrain difficulty and an increase in the kinetic friction with the ground.
Regarding the types of terrain, the proportion of stair terrains was reduced, and terrains of greater difficulty, such as open-riser stairs and grid-patterned bumps, were excluded.
Furthermore, the occurrence of ascending stairs was reduced, whereas that of descending stairs was increased.

The changes in the reward function are summarized in Table~\ref{tab:reward_functions}.
In the flat-terrain locomotion policy, the reward for velocity command tracking was increased. In addition, torque and action changes incurred higher penalties.

The input also differed from that of the rough-terrain policy: instead of one step of past observations, two exponential moving average (EMA) filters with different time constants (\SI{0.029}{\second} and \SI{0.144}{\second}) were applied to the 57-dimensional base observation, yielding a 171-dimensional input.
Unlike the rough-terrain policy, which uses single-step past observation, the flat-terrain policy captures temporal patterns such as periodic foot-lifting motions, enabled by its EMA filters that provide a longer temporal context.

As a result, the flat-terrain locomotion policy combines foot-lifting motions with wheel-based locomotion, rather than relying primarily on the wheels, as in the rough-terrain policy.
This policy achieves reduced heating during extended flat-terrain locomotion (evaluated in detail in Section~\ref{sec:exp_heat}).
Although its ability to ascend stairs and steps is to some extent inferior to that of the rough-terrain locomotion policy, its performance in descending steps is comparable.

\begin{table}[tb]
    \centering
    \caption{Reward functions and weights used in the locomotion policy training.
    Terms marked with $^*$ were introduced in~\cite{kiybib:irie_ar_2025_qrc}.
    Terms marked with $^{**}$ are newly introduced in this work.}
    \label{tab:reward_functions}
    \small
    \begin{tabular}{@{}lcc@{}}
    \toprule
    Reward term & Rough & Flat \\
    \midrule
    tracking\_lin\_vel\_var$^*$   &  1.0   &  2.0   \\
    tracking\_ang\_vel            &  0.5   &  1.0   \\
    action\_rate                  & $-0.01$  & $-0.015$ \\
    ang\_vel\_xy                  & $-0.05$  & $-0.05$  \\
    collision                     & $-1.0$   & $-1.0$   \\
    dof\_acc                      & $-2.5 \!\times\! 10^{-7}$ & $-5 \!\times\! 10^{-7}$ \\
    dof\_pos\_limits              & $-10.0$  & $-10.0$  \\
    leg\_effort\_std$^{**}$       & $-0.01$  & $-0.03$  \\
    lin\_vel\_z                   & $-2.0$   & $-2.0$   \\
    orientation                   &  0     & $-1.0$   \\
    slip$^*$                      & $-0.1$   &  0     \\
    torque\_limits                & $-0.01$  &  0     \\
    action\_curvature$^{**}$      & $-0.1$   & $-0.5$   \\
    \botrule
    \end{tabular}
\end{table}

\section{Navigation System}\label{sec:navigation}
This section describes the software system for autonomous navigation.

\subsection{Navigation Strategy}\label{sec:navigation_strategy}
The basic strategy is to perform self-localization using a pre-built 3D point cloud map and navigate along a predefined travel path.
The travel path is created by manually adjusting a sequence of waypoints automatically generated from the trajectory of a manually operated robot.
When an obstacle is detected on the path, the robot first decelerates or stops; if the obstacle is not cleared within a given time, an avoidance path is planned using A* search on an occupancy grid map, and the robot returns to the original path.
In addition, each waypoint is annotated with attributes such as the maximum speed, temporary stop, no overtaking, stair ascent, and stair descent. The obstacle avoidance modes and locomotion control are switched according to these attributes.
The switching between the flat-terrain and rough-terrain locomotion policies is also based on these attributes.

\subsection{Self-Localization}
Self-localization is performed by combining local movement estimation via lidar-inertial odometry (LIO) with cumulative error correction through point cloud matching against a prior map.
FAST-LIO2~\cite{kiybib:fastlio2} is used for LIO, and generalized iterative closest point (GICP) via small\_gicp~\cite{kiybib:small_gicp} is used for point cloud matching.
LIO runs at 10\,Hz to sequentially estimate the local movement of the robot, whereas GICP matching against the prior map is performed in parallel at approximately 1\,Hz, thereby correcting the cumulative drift of LIO and obtaining the robot pose in the map coordinate frame.

Outdoor environments often contain areas with sparse geometric features, such as open areas and straight roads, where a single scan may lack sufficient features and matching can fail.
To address this, the proposed system aligns the four most recent scan frames using local motion estimates from LIO, and combines them into an accumulated local point cloud, which is then matched against the map.
By using a point cloud covering a wider area than that in a single scan, the number of geometric features increases, thereby improving the robustness to matching failures.

\subsection{Learning-Based Odometry from Commands and IMU}
For smooth path following, local motion estimation at a higher rate than that in the Lidar odometry described in the previous subsection is required.
The motion of a legged-wheeled robot involves a combination of wheel rotation and leg movement. Moreover,
the use of pneumatic tires renders accurately detecting the ground-contact state of each foot challenging.
Therefore, a kinematics-based approach that computes the movement from the wheel rotation and leg kinematics is difficult to apply.
In this study, we trained a model via supervised learning that estimates the relative movement of the robot from a short-time history of velocity commands and used it as a real-time odometry source.

Let the velocity command at time $t$ be $\mathbf{u}_t=(v_{x,t}, v_{y,t}, \omega_t)^{\rm T}$.
The estimation model $f_\psi$ is an MLP that takes as input the velocity command history over the most recent $T$ steps,
$\{\mathbf{u}_{t-k}\}_{k=0}^{T-1}$,
and outputs the one-step differential movement in the local coordinate frame of the robot,
$\widehat{\Delta\mathbf{x}}_{t+1} = (\widehat{\Delta x}, \widehat{\Delta y})^{\rm T}$.
The network has two hidden layers of 256 units each, with a control period of 100\,Hz and history length of $T=30$.

Training data were generated from locomotion logs collected by applying random velocity commands to the robot model in the MuJoCo simulator.
Each training sample comprised a velocity command history and the corresponding relative movement at that time step.
Given $M$ collected data samples, the model parameters $\psi$ were trained by minimizing the sum of the squared prediction errors.
\[
  \psi^* =
  \mathop{\rm arg\,min}_{\psi}
  \sum_{i=1}^{M}
  \left\| f_\psi(\{\mathbf{u}_{i-k}\}_{k=0}^{T-1}) - \Delta \mathbf{x}_{i+1} \right\|^2 .
\]

During deployment on the real robot, the differential movements output by the estimation model are integrated using the yaw angle $\theta_t$ separately estimated by integrating the IMU on the onboard computer of the robot.
Specifically, the local-frame differential movement $(\widehat{\Delta x}, \widehat{\Delta y})$ is transformed into the world coordinate frame using $\theta_t$, and these are sequentially accumulated to obtain the position of the robot. The heading is obtained directly from the IMU-derived yaw angle $\theta_t$ without using the model output.
During the intervals when all velocity commands are zero, inference is skipped and the movement is set to zero to avoid accumulation of small estimation errors while stationary.

\subsection{Path-Following Control}
Classical methods such as pure pursuit are applicable to path following. However, for legged and legged-wheeled robots, the actual movement in response to a velocity command can deviate from the commanded value, leading to tracking accuracy degradation or oscillations.
We previously proposed a data-driven path-following control method for quadruped robots to address this issue~\cite{jkiybib:Irie_Robomech2021}. We adopt it in this study, as it performed well for the legged-wheeled robot as well.

Fig.~\ref{fig:path_following} illustrates the concept.
This method treats the lower-level control system, including the locomotion policy, as a black box, and learns the short-term robot movement in response to the velocity commands from the data.
The trained model is then used as a simulator on which the path-following policy $\pi_\theta$ is acquired via reinforcement learning.
A lookahead point $\mathbf{g}_t$ is set ahead of the robot on the path, and the policy $\pi_\theta$ outputs a velocity command $\mathbf{u}_t$ based on the current state and $\mathbf{g}_t$.
The robot follows the path by updating the lookahead point as it moves.

\begin{figure}[tb]
  \centering
\begin{tikzpicture}[
  >=Stealth,
  every node/.style={font=\small},
]

\coordinate (P0) at (0, 0);
\coordinate (P1) at (3.5, 0);
\coordinate (P2) at (7.5, 2.2);

\foreach \i in {0,...,7} {
  \fill[black!30] ($(P0)!\i/7!(P1)$) circle (1.5pt);
}
\foreach \i in {1,...,6} {
  \fill[black!30] ($(P1)!\i/6!(P2)$) circle (1.5pt);
}

\draw[black!50, dashed, line width=1.2pt] (P0) -- (P1) -- (P2);

\node[gray, right] at (P2) {path};

\coordinate (rob) at (2.4, 0);

\begin{scope}[shift={(rob)}, rotate=0]
  \fill[blue!30, draw=blue!60, line width=0.8pt]
    (-0.28, -0.18) -- (-0.28, 0.18) -- (0.20, 0.18) -- (0.35, 0) -- (0.20, -0.18) -- cycle;
\end{scope}

\node[blue!80] (xtlabel) at (1.5, 0.55) {$\bm{x}_t$};
\draw[->, blue!70, thin] (xtlabel) -- ($(rob) + (-0.1, 0.15)$);

\coordinate (la) at (4.9, 0.77);
\fill[green!50!black] (la) +(-0.08,-0.08) rectangle +(0.08,0.08);

\draw[green!60!black, dotted, line width=1pt] (rob) -- (la);

\node[green!50!black] (lalabel) at (5.8, -0.2) {look-ahead $\bm{g}_t$};
\draw[->, green!50!black, thin] (lalabel) -- ($(la) + (0.1, -0.1)$);

\coordinate (next) at (3.95, 0.32);

\begin{scope}[shift={(next)}, rotate=25]
  \fill[blue!15, draw=blue!30, line width=0.6pt, dashed]
    (-0.28, -0.18) -- (-0.28, 0.18) -- (0.20, 0.18) -- (0.35, 0) -- (0.20, -0.18) -- cycle;
\end{scope}

\node[blue!60] (xnextlabel) at (4.3, 1.2) {$\bm{x}_{t+1}$};
\draw[->, blue!50, thin] (xnextlabel) -- ($(next) + (0.05, 0.12)$);

\coordinate (umid) at ($(rob)!0.5!(next)$);
\draw[->, orange!80, line width=1.5pt] ($(rob) + (0.2, 0.05)$) -- ($(next) + (-0.15, -0.05)$);
\node[orange!80!black, above left=-1pt and 1pt, font=\footnotesize] at (umid) {$\bm{u}_t$};

\node[draw=orange!70, fill=orange!8, rounded corners=4pt,
      minimum width=3.2cm, minimum height=0.7cm,
      font=\normalsize, text=orange!80!black]
  (policy) at (3.2, -1.4)
  {$\bm{u}_t = \pi_\theta(\bm{s}_t,\, \bm{g}_t)$};

\draw[orange!60, dashed, line width=0.8pt] (policy.north) -- ($(umid) + (0, -0.1)$);

\end{tikzpicture}
  \caption{Conceptual diagram of the path-following control. At each time step, a look-ahead point $\mathbf{g}_t$ is set ahead on the path, and the policy $\pi_\theta$ outputs a velocity command $\mathbf{u}_t$ based on the current state and $\mathbf{g}_t$.
  The look-ahead point is updated as the robot moves to follow the path.}
  \label{fig:path_following}
\end{figure}

Details of the movement model, path-following policy training, and a comparison with pure pursuit are provided in Appendix~B.

\section{Experiments}\label{sec:experiment}

\subsection{Locomotion Performance Evaluation}

Here, we present basic evaluations of the flat-terrain and rough-terrain locomotion policies through simulations and experiments on the physical robot.

\subsubsection{Evaluation in Simulated Environments}

In the simulations, we evaluated the difference in the foot-lifting behavior between the two policies, tracking performance for commanded velocities, and traversability on several types of terrain.

\begin{figure*}[tb]
  \centering
  \includegraphics[width=\textwidth]{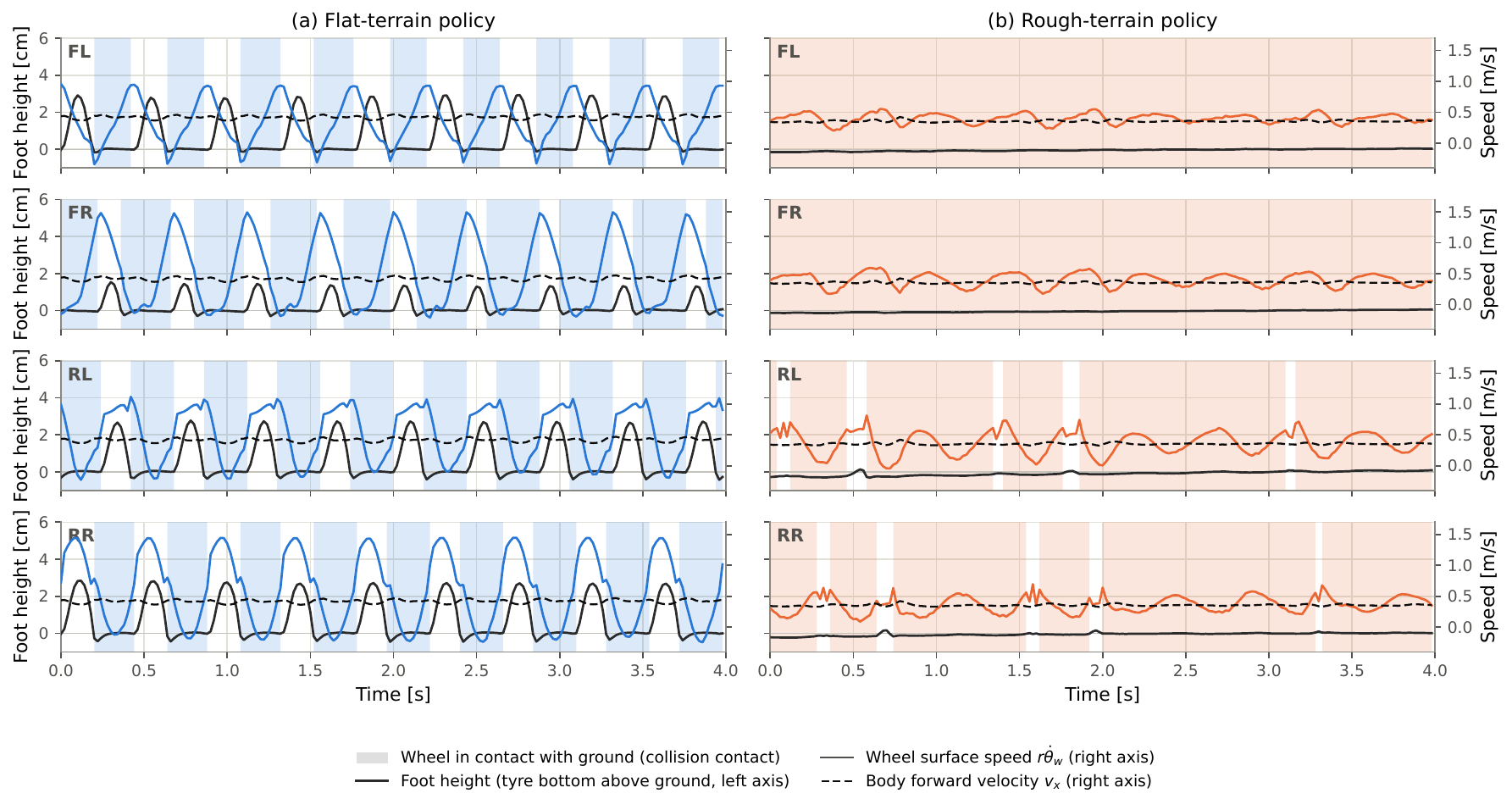}
  \caption{Wheel contact pattern, foot height, and wheel surface speed of each leg during 4 s of forward travel at \SI{0.4}{\meter\per\second} on flat ground.
  Shaded regions indicate ground contact of the wheel; the solid black line is the foot height (bottom of the tire above the ground, left axis); the colored line is the wheel surface speed $r\dot{\theta}_w$ and the dashed black line is the body forward velocity (right axis).
  The rough-terrain policy keeps the wheels in contact with the ground almost all the time, and the wheel surface speed is close to the body speed.
  In contrast, the flat-terrain policy lifts the legs alternately with a period of about \SI{0.5}{\second}. The wheel surface speed is increased while the legs are in contact with the ground, resulting in combined wheel-leg locomotion.}
  \label{fig:sim_gait}
\end{figure*}

First, we describe the difference in foot-lifting behavior.
Fig.~\ref{fig:sim_gait} shows the foot-lifting pattern and wheel surface speed of each wheel while the robot moves forward on a flat ground at \SI{0.4}{\meter\per\second}. The wheel surface speed is the tangential speed of the wheel rim, $r\dot{\theta}_w$.
The rough-terrain policy kept all four wheels in contact with the ground almost consistently. The wheel surface speed remained close to the body speed, indicating that the robot moved by rotating the wheels.
In contrast, the flat-terrain policy lifted the four legs alternately with a period of approximately \SI{0.5}{\second}.
The duty factor, defined as the fraction of time during which each wheel is in contact with the ground averaged over the four wheels, was close to 1 for the rough-terrain policy and approximately 0.6 for the flat-terrain policy.
The wheel surface speed increased during the stance phase and began to decrease at the transition to the swing phase.
These results show that the flat-terrain policy combines active wheel rotation with periodic foot-lifting motions.

\begin{figure*}[tb]
  \centering
  \includegraphics[width=\textwidth]{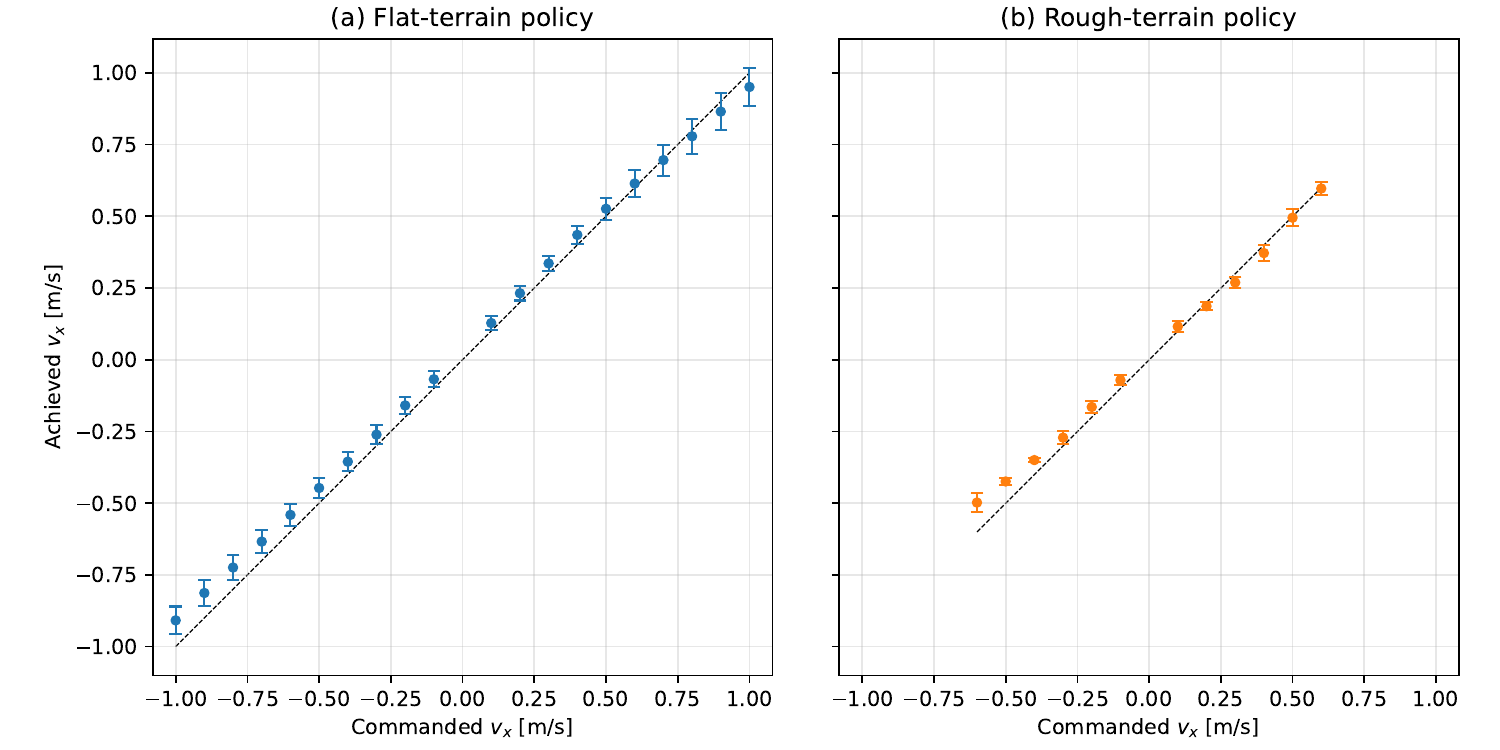}
  \caption{Result of velocity tracking experiments using the flat-terrain and rough-terrain policies in simulation.
  A \SI{10}{\second} run was performed for each commanded forward and backward velocity, and the mean and standard deviation of the achieved velocity are plotted.
  Note that the velocity ranges differ between the two policies because the command ranges used during training differ.}
  \label{fig:sim_tracking}
\end{figure*}

Fig.~\ref{fig:sim_tracking} shows the mean and standard deviation of the achieved velocity when forward and backward velocity commands within the training ranges were given on flat ground.
For both policies, the achieved velocity was approximately proportional to the commanded velocity, and the difference between the mean achieved velocity and the command was at most approximately \SI{0.1}{\meter\per\second} regardless of the difference in foot-lifting behavior.
Both policies tended to fall slightly short of the command at larger command magnitudes, particularly when moving backward.

\begin{figure}[tb]
  \centering
  \includegraphics[width=\linewidth]{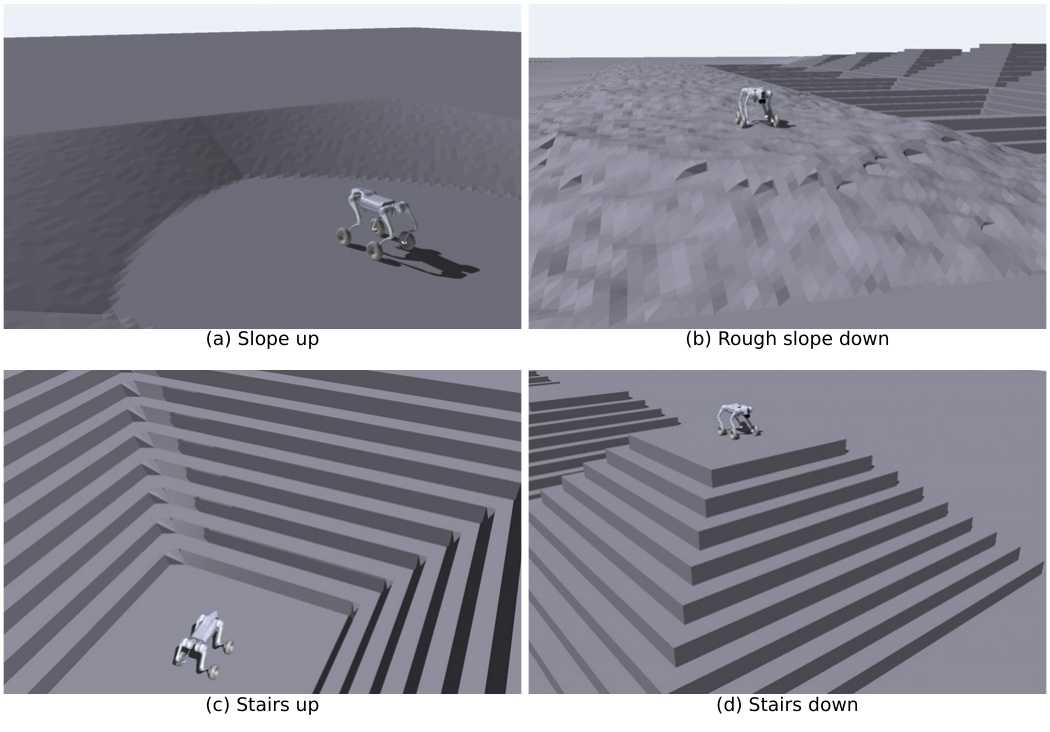}
  \caption{Simulated terrains used for the traversal evaluation: (a) uphill slope, (b) downhill rough slope, (c) stairs up, (d) stairs down.}
  \label{fig:sim_terrain_layout}
\end{figure}

Using the four types of terrain shown in Fig.~\ref{fig:sim_terrain_layout}, we evaluated the traversal success rates of both policies on slopes and stairs.
The four terrain types are an uphill slope, a downhill rough slope, upward stairs, and downward stairs.
Each terrain is an \SI{8}{\meter} square with a \SI{3}{\meter} square flat platform at the center, and the steps or slope are arranged outward from the platform.
On the downhill rough slope, random height variations of $\pm\SI{5}{\centi\meter}$ at \SI{20}{\centi\meter} intervals were superimposed.

Each terrain type was generated at 11 difficulty levels, in which the slope angle or step height increased with the level.
The step height of the stairs ranged from \SI{5}{\centi\meter} to \SI{25}{\centi\meter}, with a constant tread depth of \SI{30}{\centi\meter}.
The slope angle ranged from \SI{0}{\degree} to \SI{20}{\degree}.

Sixty-four robots were placed on each terrain. The initial position of each robot was sampled uniformly within $\pm\SI{0.5}{\meter}$ of the center of the flat platform, and the initial heading was sampled uniformly within $\pm\SI{20}{\degree}$ of the direction perpendicular to the terrain to be traversed. Each robot was driven for \SI{20}{\second} with a forward velocity command of \SI{0.4}{\meter\per\second} and a yaw-rate command computed by feedback to maintain the initial heading.
A trial was considered successful if the robot traveled at least \SI{3.7}{\meter} from the start without falling; the success rate was computed for each difficulty level.
In addition, to evaluate the effect of the DRL-based leg control, we implemented a wheel-only locomotion mode with fixed legs and compared its performance with that of the two learned policies. In this condition, the leg joints were held at the default posture by stiff proportional-derivative (PD) control and only the wheels were velocity-controlled. Note that this condition was not intended to mimic the manufacturer-provided controller.

\begin{figure}[tb]
  \centering
  \includegraphics[width=\linewidth]{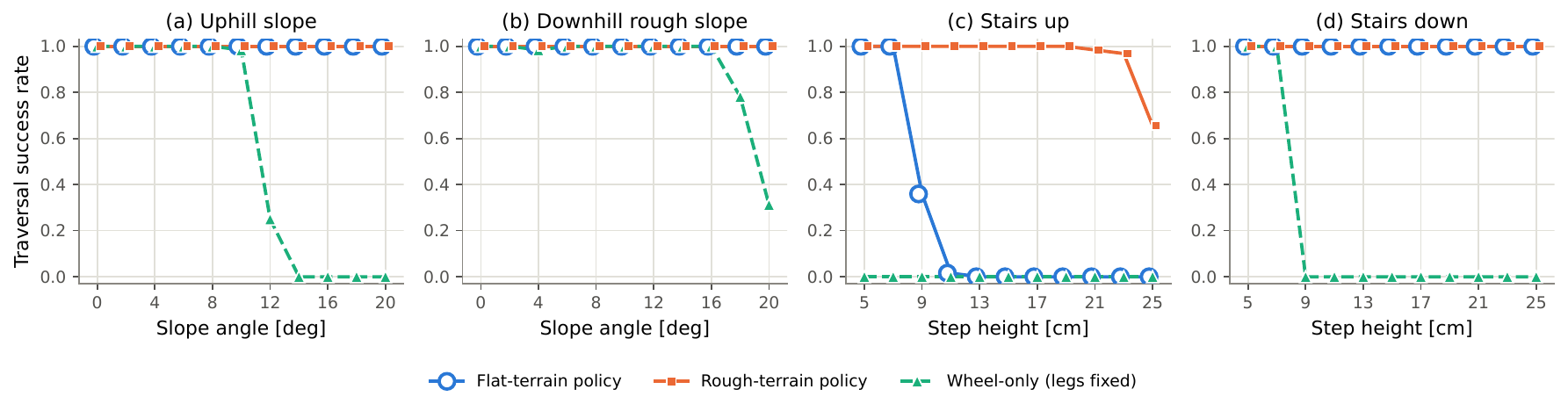}
  \caption{Terrain traversal success rate at \SI{0.4}{\meter\per\second} (64 robots per terrain, \SI{20}{\second}): (a) uphill slope, (b) downhill rough slope, (c) stairs up, (d) stairs down.}
  \label{fig:sim_terrain}
\end{figure}

Fig.~\ref{fig:sim_terrain} shows the traversal success rate for each terrain and difficulty level.
On the uphill slope, downhill rough slope, and downward stairs, both proposed policies traversed all difficulty levels.
On the upward stairs, the flat-terrain policy achieved a success rate of 90\% or higher up to a step height of \SI{7}{\centi\meter}, and the rough-terrain policy up to \SI{23}{\centi\meter}. Thus, the rough-terrain policy showed higher traversability.
The wheel-only locomotion with fixed legs traversed the uphill slope up to \SI{10}{\degree} and the downward stairs up to \SI{7}{\centi\meter}. However, beyond these levels, the robot fell. It could not navigate the upward stairs at all.

These results confirm that the two locomotion policies obtained in this study track the velocity commands within their training ranges. They also confirm that the leg motions acquired through DRL contribute to the robustness of traversing slopes and steps.
Based on the difference in the step-traversal characteristics between the two policies, the rough-terrain policy was used on the stair sections and flat-terrain policy elsewhere in the navigation experiments in this study.

\subsubsection{Evaluation on the Real Robot}

\begin{figure}[tb]
  \centering
  \newcommand{\gaitframe}[1]{\stepcounter{frameno}%
    \labelframe{\theframeno}{width=0.19\linewidth}{#1}}
  \tabcolsep=1pt
  \renewcommand{\arraystretch}{0.5}
  \begin{tabular}{@{}ccccc@{}}
    \setcounter{frameno}{0}%
    \gaitframe{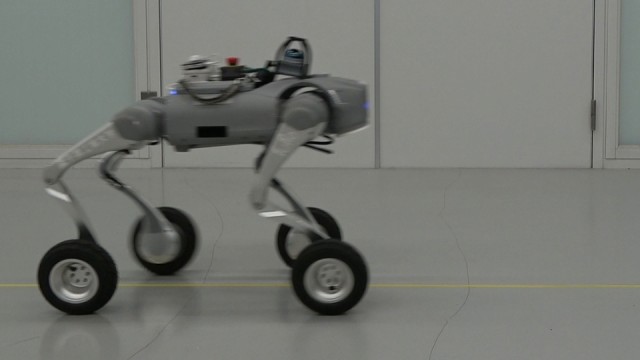} &
    \gaitframe{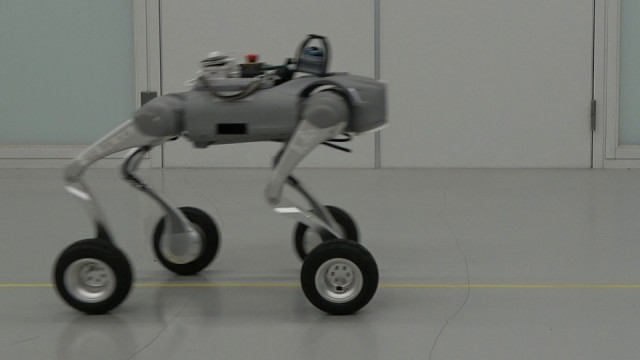} &
    \gaitframe{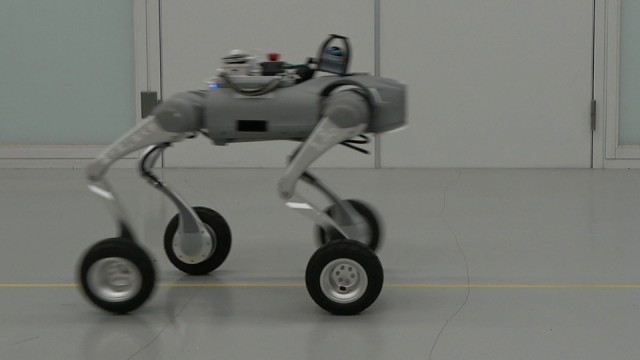} &
    \gaitframe{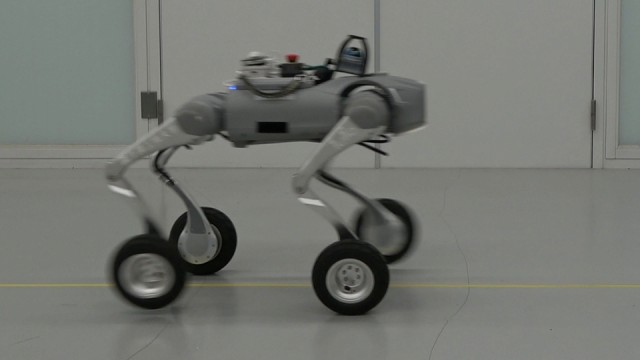} &
    \gaitframe{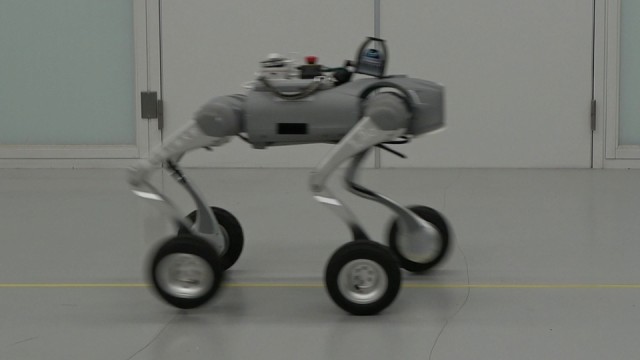} \\
    \gaitframe{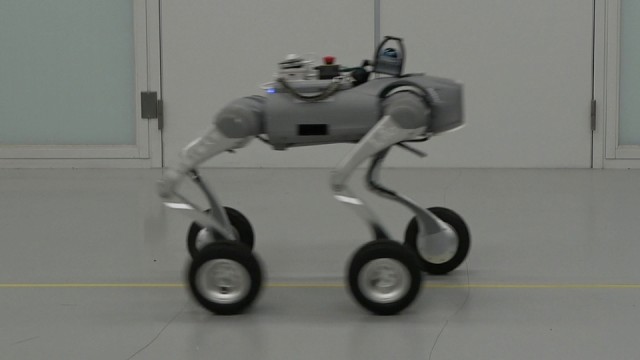} &
    \gaitframe{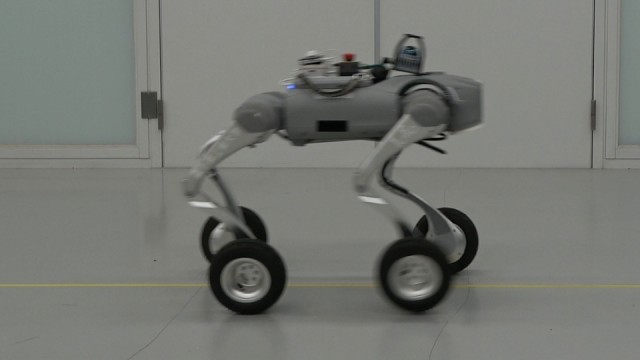} &
    \gaitframe{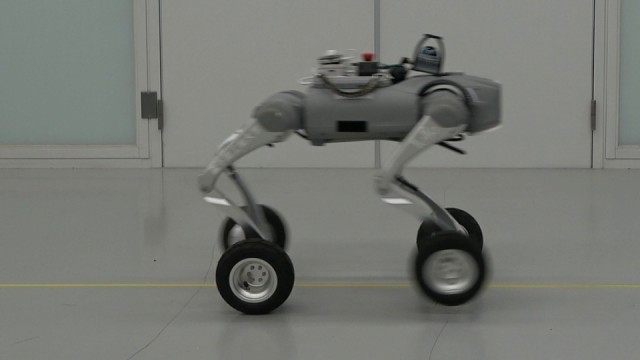} &
    \gaitframe{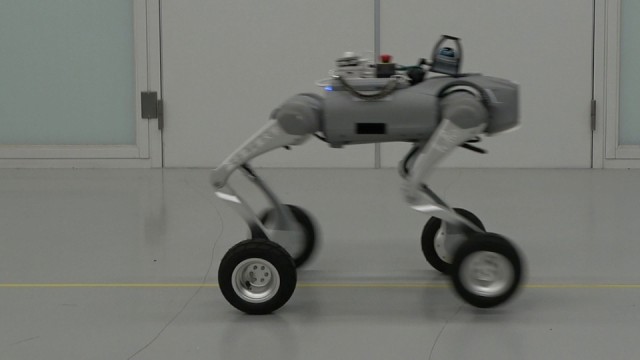} &
    \gaitframe{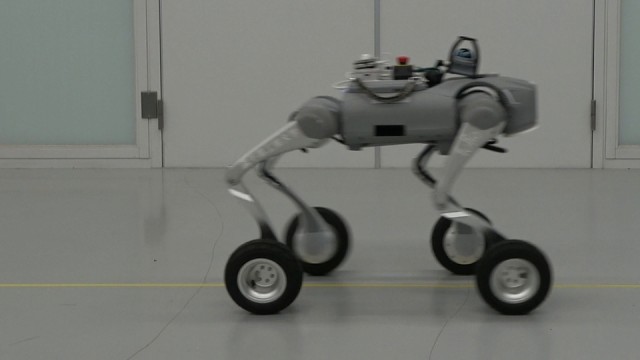} \\
    \multicolumn{5}{c}{(a) Flat-terrain policy} \\[6pt]
    \setcounter{frameno}{0}%
    \gaitframe{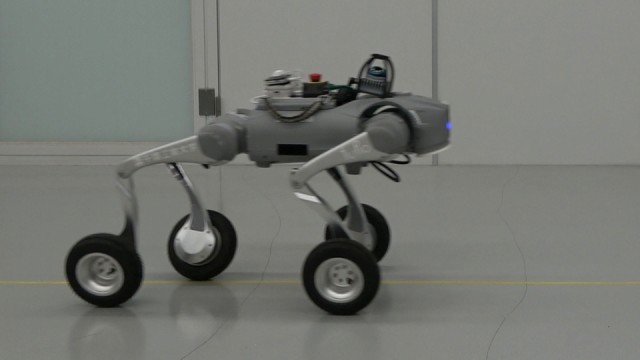} &
    \gaitframe{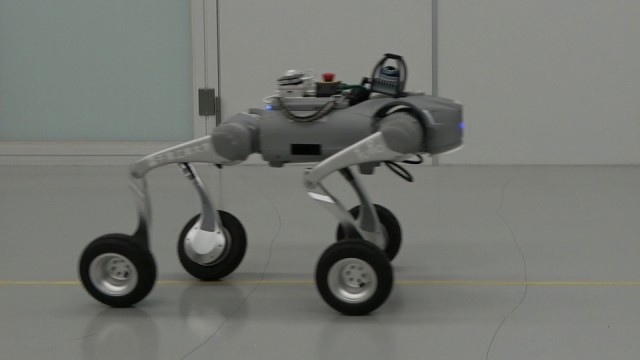} &
    \gaitframe{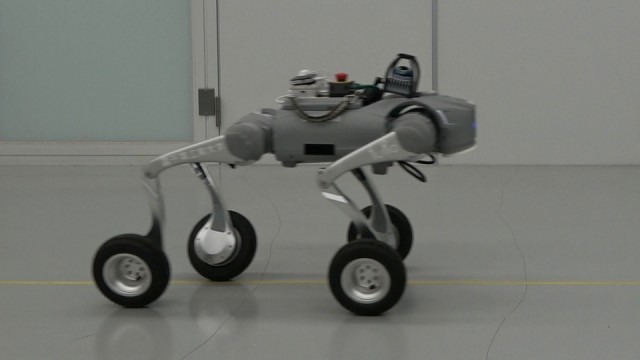} &
    \gaitframe{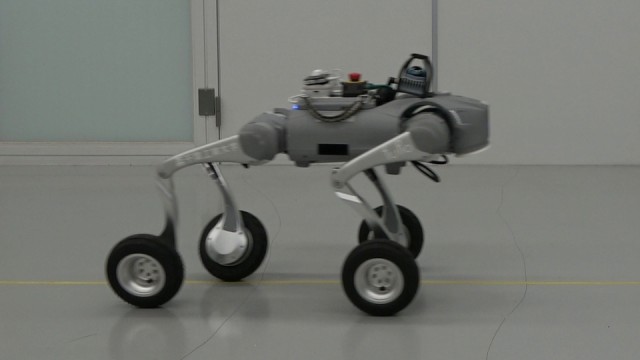} &
    \gaitframe{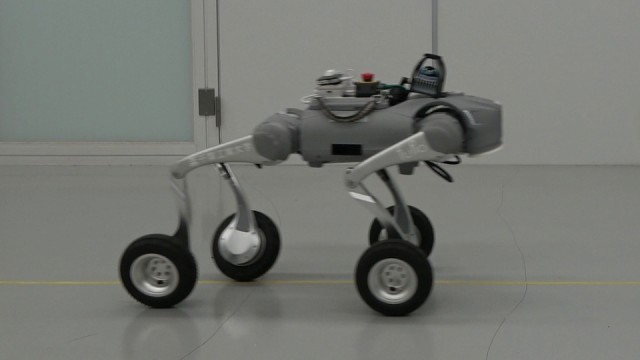} \\
    \gaitframe{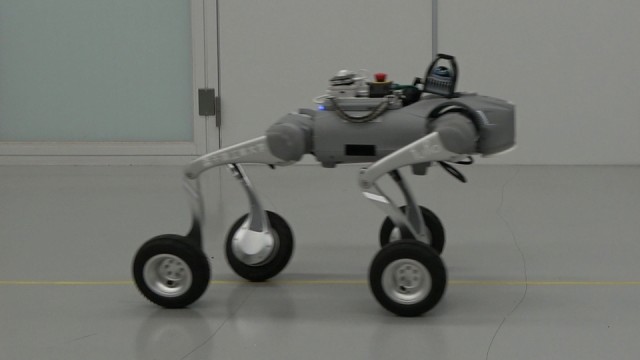} &
    \gaitframe{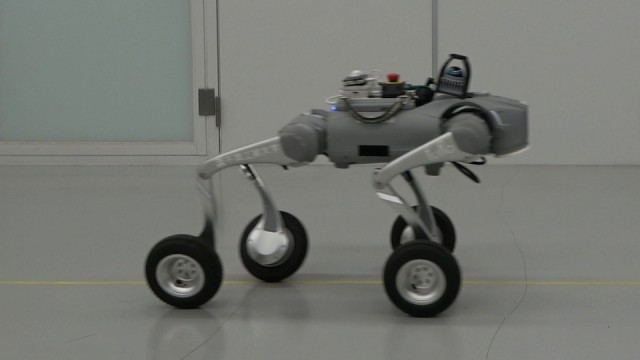} &
    \gaitframe{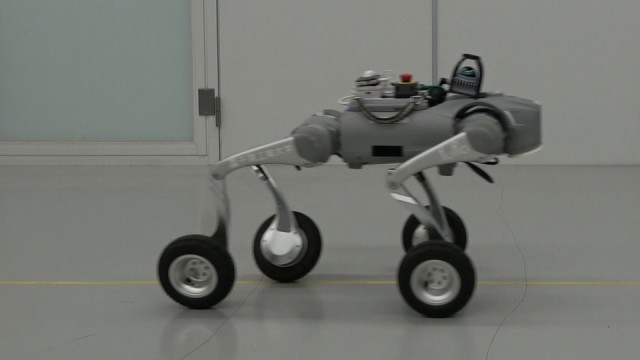} &
    \gaitframe{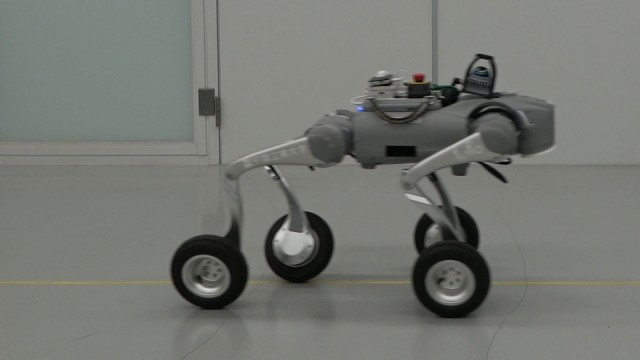} &
    \gaitframe{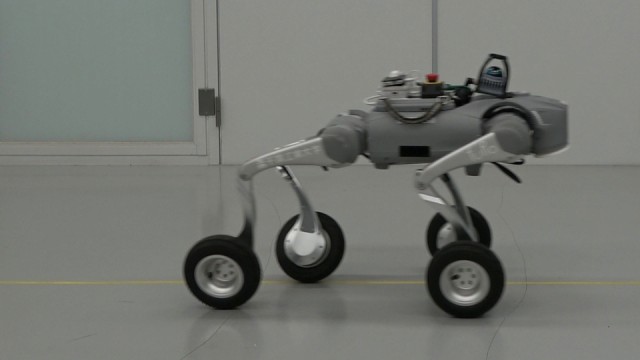} \\
    \multicolumn{5}{c}{(b) Rough-terrain policy}
  \end{tabular}
  \caption{Locomotion gaits of the two policies. (a) The flat-terrain policy combines periodic foot-lifting motions with wheel rotations. (b) The rough-terrain policy uses primarily wheel rotations without foot-lifting.}
  \label{fig:ctrl_gait}
\end{figure}

We tested both policies on a flat ground with the real robot.
Fig.~\ref{fig:ctrl_gait} shows the motion of the robot under each locomotion policy when given a velocity command of \SI{1.0}{\meter\per\second}.
The flat-terrain policy exhibits periodic foot lifting, whereas the rough-terrain policy keeps all four wheels in contact with the ground and moves by rotating the wheels.
Thus, both policies reproduced on the real robot the gaits observed in the simulation.

\begin{figure}
  \centering
  \newcommand{\stepframe}[1]{\stepcounter{frameno}%
    \labelframe{\theframeno}{width=0.24\linewidth}{#1}}
  \tabcolsep=1pt
  \renewcommand{\arraystretch}{0.5}
  \begin{tabular}{@{}cccc@{}}
    \setcounter{frameno}{0}%
    \stepframe{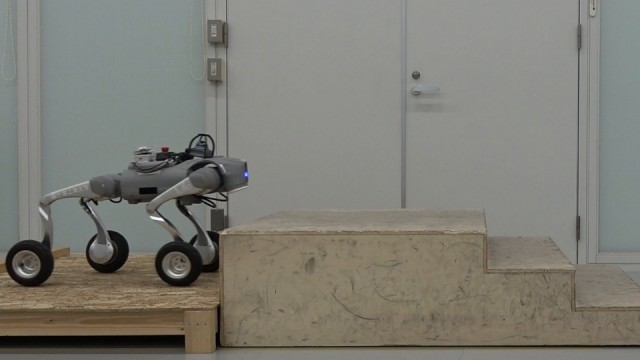} &
    \stepframe{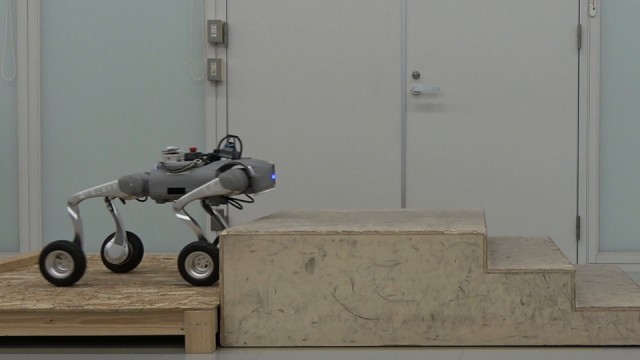} &
    \stepframe{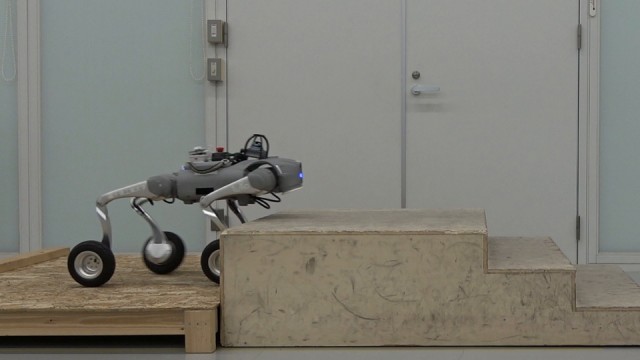} &
    \stepframe{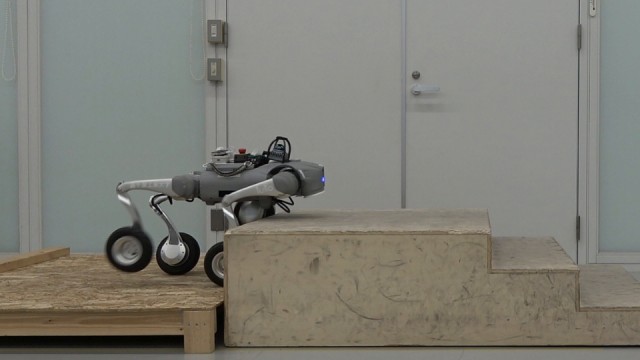} \\
    \stepframe{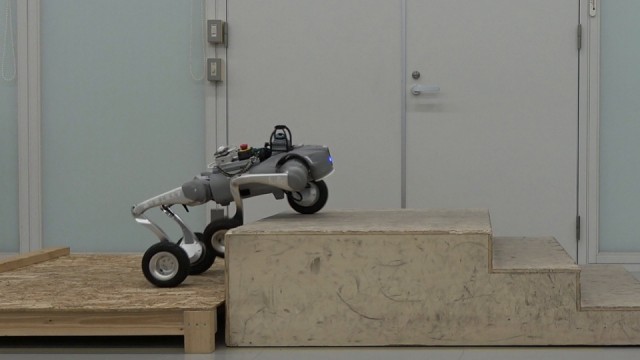} &
    \stepframe{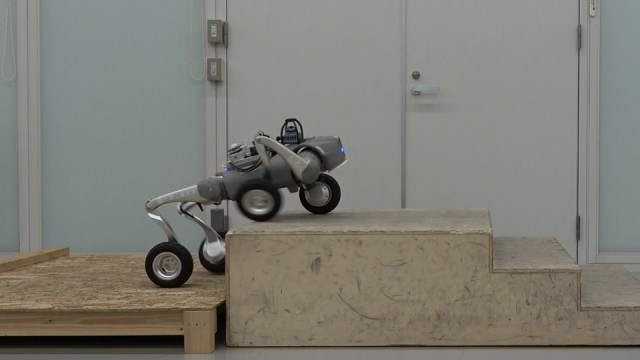} &
    \stepframe{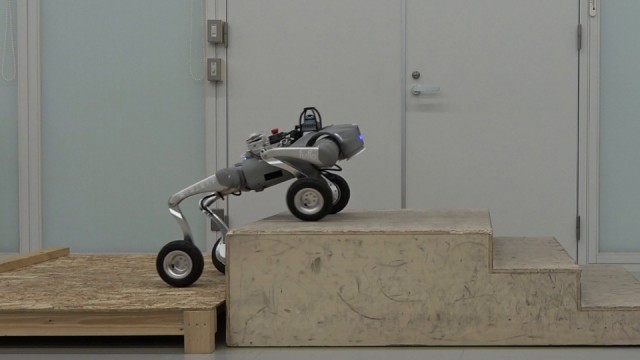} &
    \stepframe{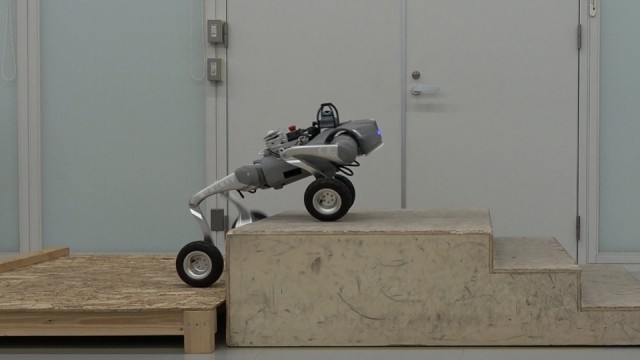} \\
    \stepframe{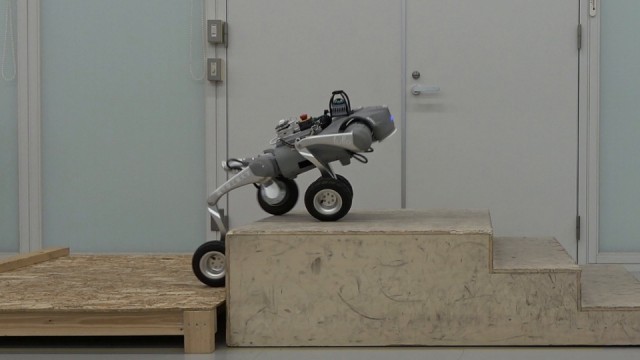} &
    \stepframe{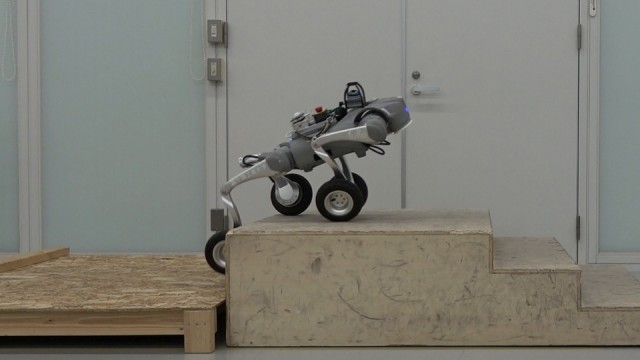} &
    \stepframe{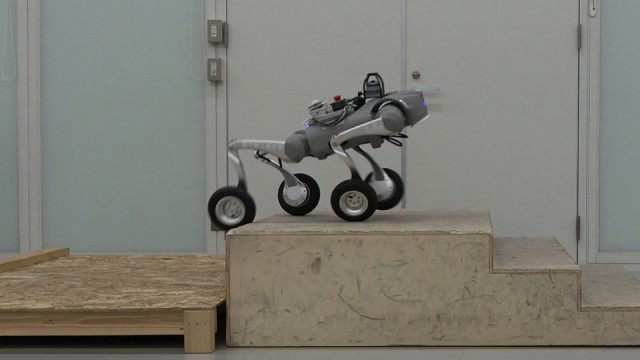} &
    \stepframe{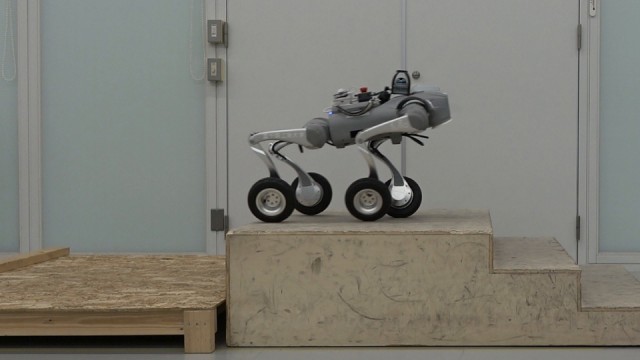} \\
    \multicolumn{4}{c}{(a) Ascending and descending a single step (\SI{24}{\centi\meter} height)} \\[6pt]
    \setcounter{frameno}{0}%
    \stepframe{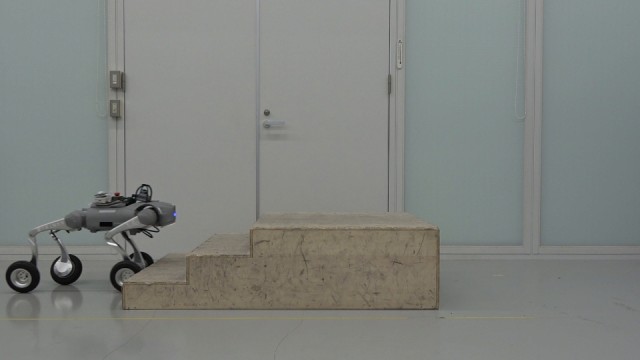} &
    \stepframe{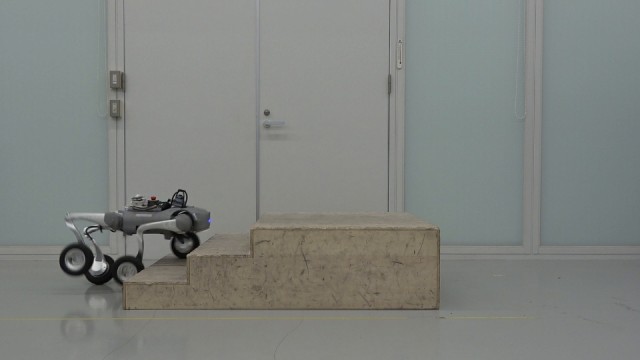} &
    \stepframe{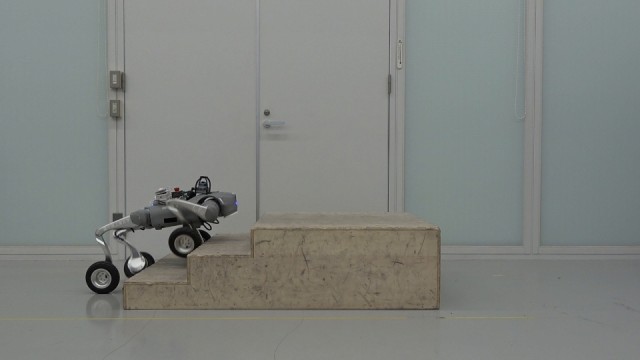} &
    \stepframe{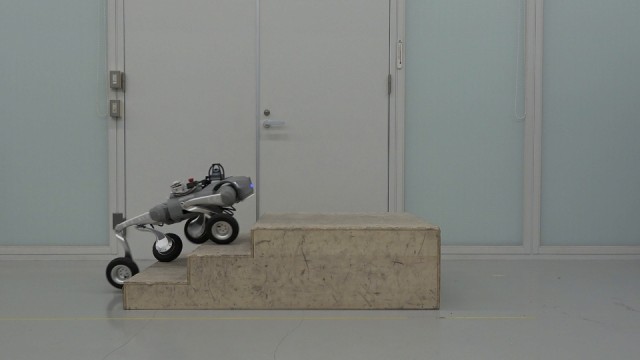} \\
    \stepframe{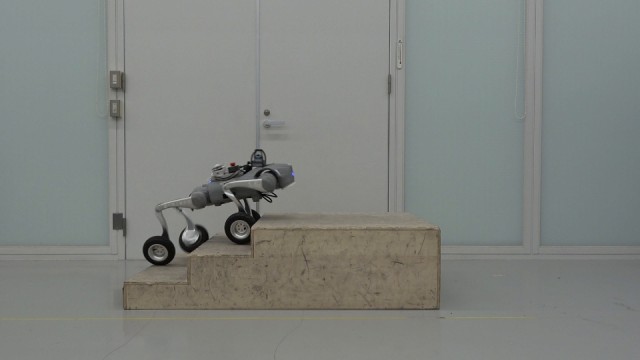} &
    \stepframe{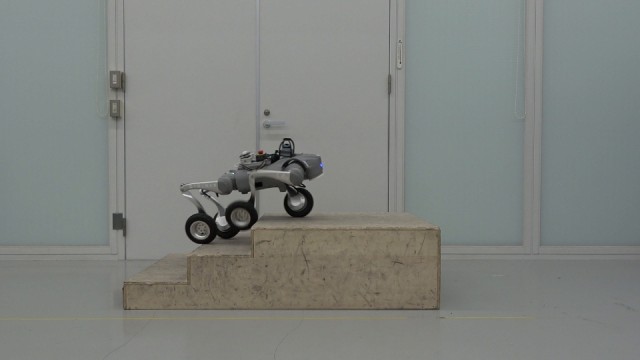} &
    \stepframe{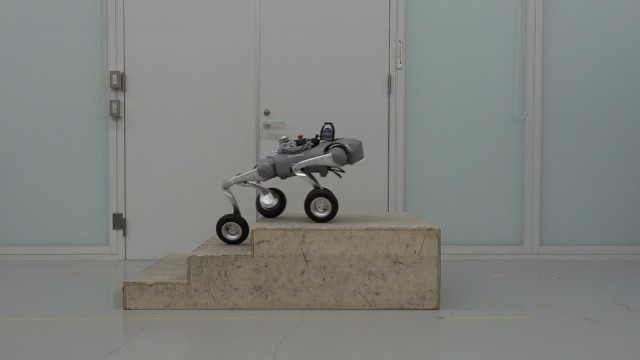} &
    \stepframe{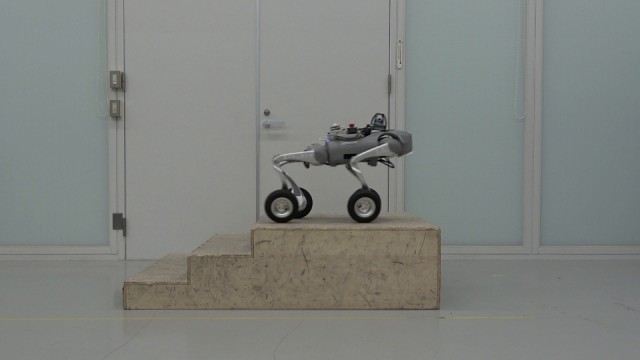} \\
    \stepframe{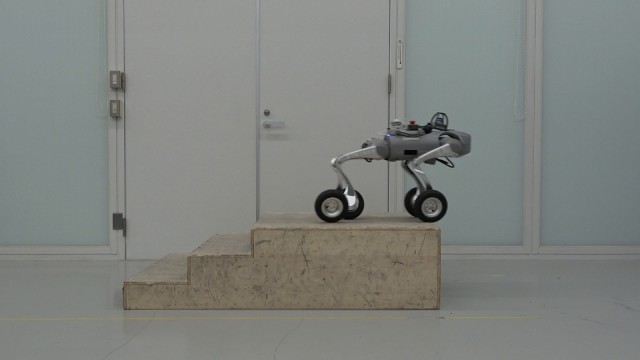} &
    \stepframe{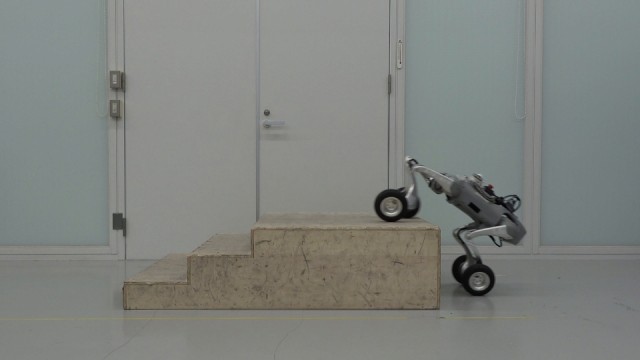} &
    \stepframe{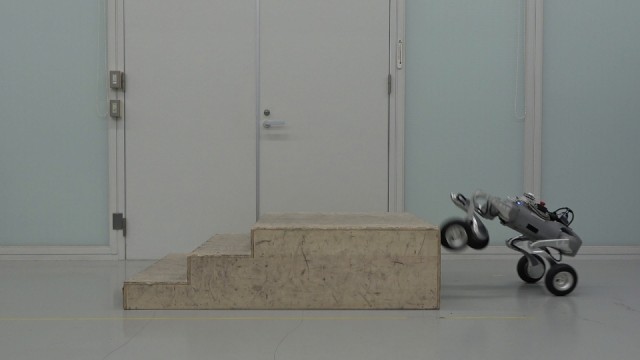} &
    \stepframe{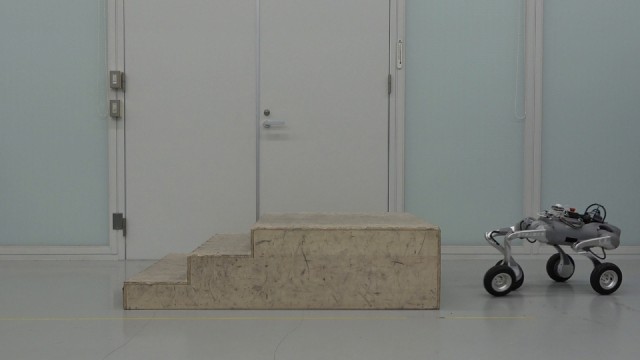} \\
    \multicolumn{4}{c}{(b) Ascending and descending a staircase (step height: \SI{13}{\centi\meter})}
  \end{tabular}
  \caption{Step and staircase traversal using the rough-terrain locomotion policy.}
  \label{fig:step_climb}
\end{figure}

Next, we measured the step-traversal capability of each policy.
The rough-terrain policy enabled the robot to ascend or descend a single step up to a height of \SI{24}{\centi\meter} (Fig.~\ref{fig:step_climb}(a)).
The robot could also ascend or descend a staircase with a step height of \SI{13}{\centi\meter} (Fig.~\ref{fig:step_climb}(b)).

The flat-terrain policy enabled traversing a step height of \SI{10}{\centi\meter}.
However, the robot could not ascend a staircase with a step height of \SI{13}{\centi\meter}, although it could descend the same staircase.
These results show traversability similar to that observed in the simulation.

\subsection{Field Experiments at the Tsukuba Challenge 2025}\label{sec:Tsukuba_Challenge}
We evaluated the effectiveness of the proposed approach through field experiments at the Tsukuba Challenge 2025.

\subsubsection{Experimental Conditions and Course}

The Tsukuba Challenge is an annual field experiment held in Tsukuba City, Ibaraki Prefecture, Japan, for verifying autonomous mobile-robot navigation in real-world environments~\cite{yuta2018,hara2020}.
Rather than being a competitive race, the event aims to foster technology development and information sharing on robotic systems operating in real-world environments~\cite{morales2008}.
The event consists of multiple experimental run days, and a final demonstration run on the last day.
All robots are required to have an emergency stop button; the accompanying operator must immediately stop the robot if a hazardous situation is perceived.
At crosswalks and signalized intersections, the robot is required to stop autonomously, and the operator confirms safety before the robot resumes navigation.

\begin{figure}
  \centering
  \includegraphics[width=\linewidth]{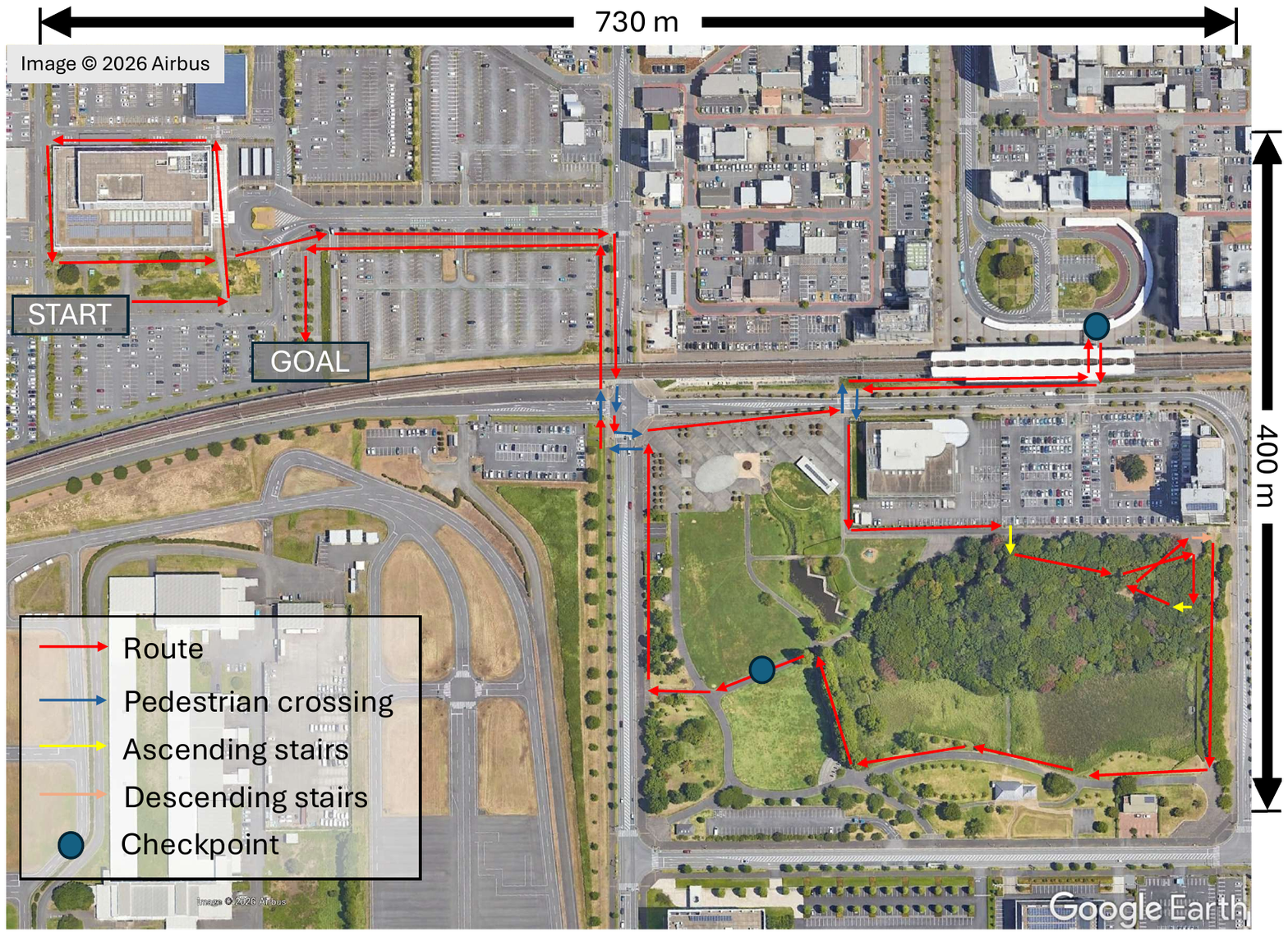}
  \caption{Satellite image of the Tsukuba Challenge 2025 area with the navigation route overlaid. The overlaid route is the approximately \SI{2.8}{\kilo\meter} course used in our experiments, starting from Tsukuba City Hall, passing through checkpoints at a train station and a park, and returning to the city hall.}
  \label{fig:TC_MAP}
\end{figure}

The course for the Tsukuba Challenge 2025 started at Tsukuba City Hall, passed through two checkpoints located at a train station and a park, and returned to the city hall, covering a distance of approximately \SI{2.8}{\kilo\meter}.
Fig.~\ref{fig:TC_MAP} shows the course.
No traffic restrictions are imposed during the runs, and ordinary pedestrians share the walkways.
The robot must therefore detect obstacles and pedestrians on the course and take appropriate actions such as stopping or avoidance.
In addition, road cones are randomly placed as static obstacles in certain sections of the course.

The navigable area included a park.
Within the park, participating teams were free to choose their own route.
We chose a route that included staircases to test the step-traversal capability of the robot.

The course contained three sets of stairs: the first was a three-step staircase with a step height of \SI{13}{\centi\meter}. The second was an 18-step staircase with a landing, and the third was an eight-step staircase with a step height of approximately \SI{20}{\centi\meter}, located in an unpaved area.
This unpaved area was situated in the forested section shown in the lower right of Fig.~\ref{fig:TC_MAP}, where the ground soil was covered with fallen leaves and twigs.

\subsubsection{Prior Map Construction}
\begin{figure}[tb]
    \centering
    \begin{tabular}{cc}
        \includegraphics[height=4cm]{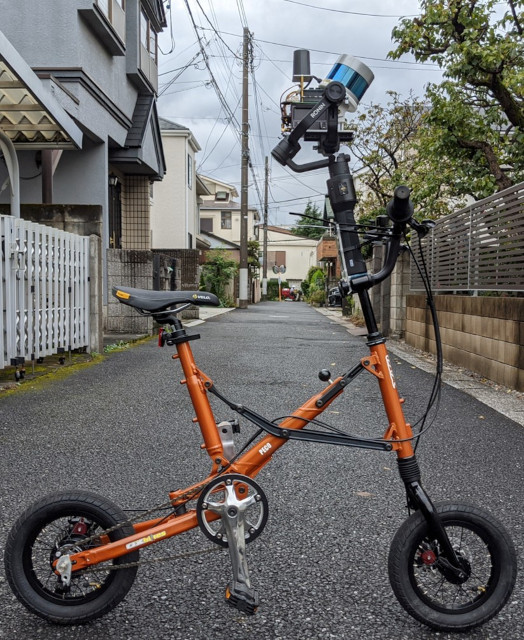} &
        \includegraphics[height=4cm]{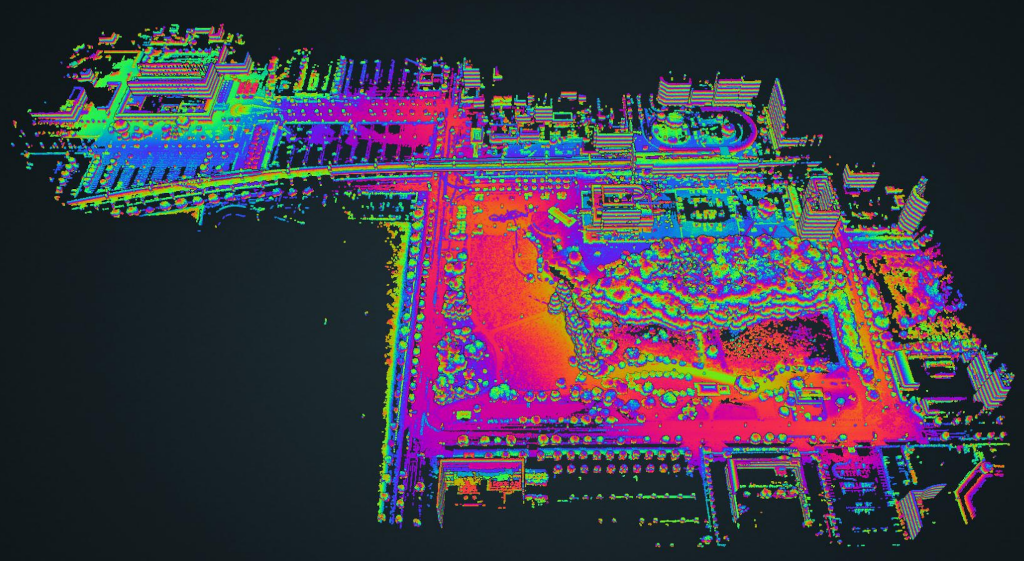} \\
        (a) Sensor-equipped bicycle & (b) Constructed 3D point cloud map
    \end{tabular}
    \caption{(a) Bicycle equipped with a Lidar, an IMU, and a GNSS receiver, used for data collection. (b) 3D point cloud map constructed by loosely coupling Lidar SLAM with GNSS data.}
    \label{fig:map_construction}
\end{figure}

The prior map used for navigation and localization was constructed by collecting data using a sensor-equipped bicycle, as shown in Fig.~\ref{fig:map_construction}(a).
The bicycle was equipped with a lidar (Hesai XT32 M2X), an IMU (Analog Devices ADIS16505-2), and a GNSS receiver (Septentrio AstRx).
Data were collected by riding the bicycle throughout the area designated as navigable for the Tsukuba Challenge 2025.
A three-dimensional point cloud map with global geospatial coordinates was generated by loosely coupling lidar SLAM with GNSS data.
GLIM~\cite{glim}, an open-source lidar SLAM framework, was used for this purpose.
The constructed map is shown in Fig.~\ref{fig:map_construction}(b).
The resulting three-dimensional point cloud map is publicly available as map\_tc25\_gnss\_furo in the Tsukuba Challenge dataset repository~\cite{tc-datasets}.

\subsubsection{Preliminary Experimental Run}

Prior to developing our locomotion control system, we conducted a teleoperated test run on October 4 to assess whether the control provided by the manufacturer could sustain locomotion over the full course.

During this run, the locomotion control system stopped with an error at a point \SI{1.4}{\kilo\meter} from the starting point owing to the temperature of the hip motor of the rear-left leg rising to \SI{86}{\degreeCelsius}.
This triggered a thermal shutdown to protect the hardware.

This experiment confirmed that completing the entire course of the Tsukuba Challenge was difficult with the control provided by the manufacturer, and that development of a control method with reduced heating was necessary.
The detailed temperature histories are described in Section~\ref{sec:tc_temp} together with those of the runs with the proposed system.

\subsubsection{Results}

\begin{table}[tb]
    \centering
    \caption{Run records of autonomous navigation experiments at the Tsukuba Challenge 2025. Duration is in hours:minutes.}
    \label{tab:TC_EXP_RES}
    \small
    \begin{tabular}{@{}lccccl@{}}
    \toprule
    \# & Date & Start & End & Duration & Intervention \\
    \midrule
    1 & 12/05 & 10:33 & 11:38 & 1:05 & None  \\
    2 & 12/05 & 13:11 & 14:20 & 1:09 & Unexpected obstacle removed \\
    3 & 12/05 & 14:42 & 15:52 & 1:10 & E-stop for oncoming vehicle \\
    4 & 12/06 & 10:16 & 11:20 & 1:04 & None  \\
    5 & 12/06 & 12:55 & 14:01 & 1:06 & None  \\
    6 & 12/07 & 11:58 & 13:02 & 1:04 & None  \\
    \botrule
    \end{tabular}
    \normalsize
\end{table}

Table~\ref{tab:TC_EXP_RES} shows the results of the autonomous navigation experiments conducted over three days from December 5 to 7.
The robot completed the course from start to endpoint in all six runs.
Each run involved continuous locomotion exceeding one hour, yet the locomotion control system did not overheat and shut down, confirming the effectiveness of the heat reduction measures.

Of the six runs, two runs on December 5 (runs \#2 and \#3) required human intervention.
The intervention in run \#2 occurred in a zone near a road crossing where obstacle avoidance had been intentionally disabled to prevent the robot from deviating onto the roadway.
A road cone had been placed outside its designated area to block the path in this zone, and the accompanying operator removed it to allow the run to resume.

The intervention in run \#3 was to avoid a parked vehicle on the road.
A parked truck was present on the course, and avoiding it required the robot to move into the oncoming lane.
However, this location was in a blind spot for oncoming vehicles. Our robot does not have the capability to detect and safely avoid fast-moving objects such as automobiles.
Therefore, to prevent a potential collision with oncoming traffic, the accompanying operator pressed the emergency stop button to halt the robot.
After the oncoming vehicle had passed, the emergency stop was released and the run was resumed.

\begin{figure*}[tb]
  {
  \tabcolsep=0.5mm
  \renewcommand{\arraystretch}{0.5}
  \begin{tabular}{cccc}
    \includegraphics[width=0.24\linewidth]{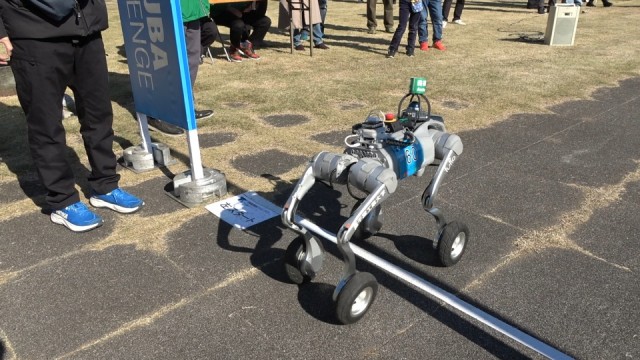}&
    \includegraphics[width=0.24\linewidth]{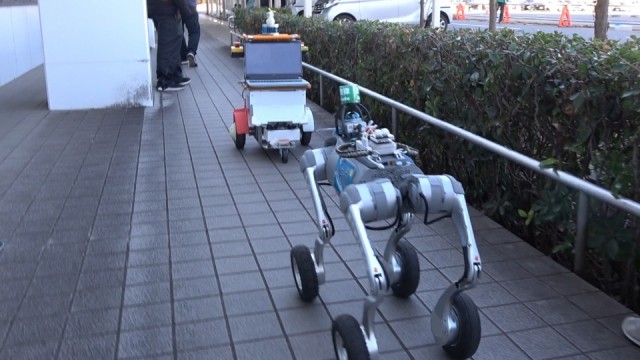}&
    \includegraphics[width=0.24\linewidth]{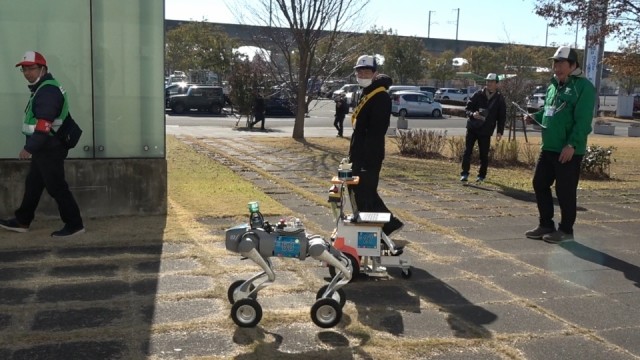}&
    \includegraphics[width=0.24\linewidth]{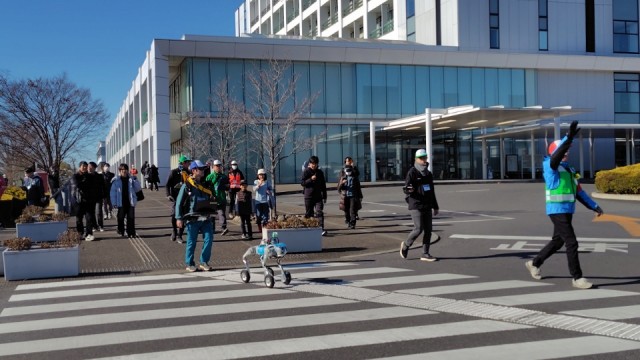}
    \\
    (a) & (b) & (c) & (d) \\[2mm]
    \includegraphics[width=0.24\linewidth]{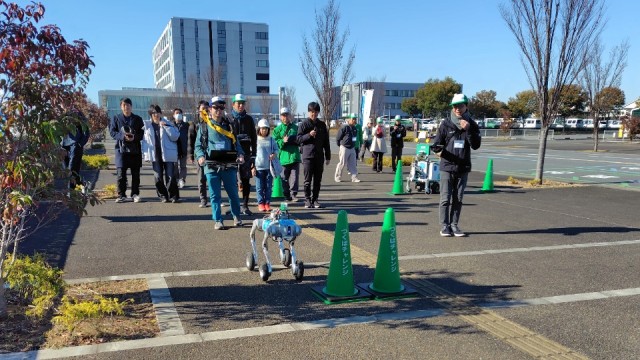}&
    \includegraphics[width=0.24\linewidth]{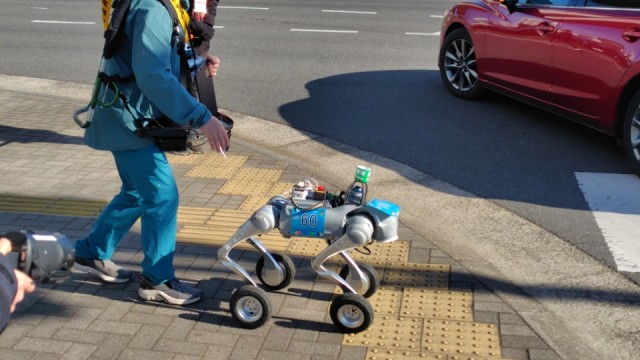}&
    \includegraphics[width=0.24\linewidth]{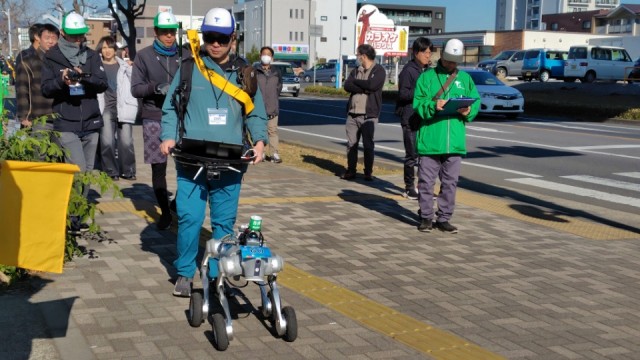}&
    \includegraphics[width=0.24\linewidth]{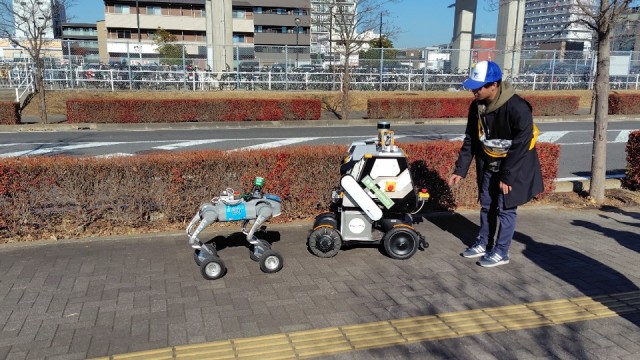}
    \\
    (e) & (f) & (g) & (h) \\[2mm]
    \includegraphics[width=0.24\linewidth]{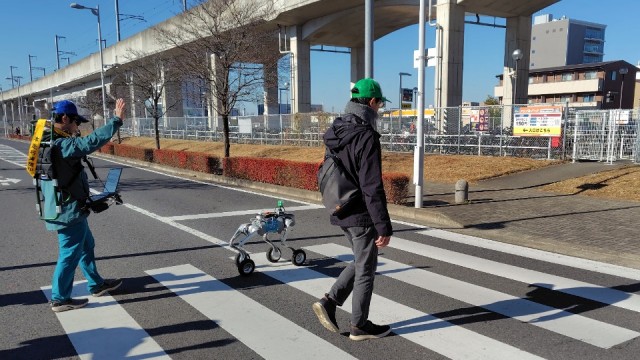}&
    \includegraphics[width=0.24\linewidth]{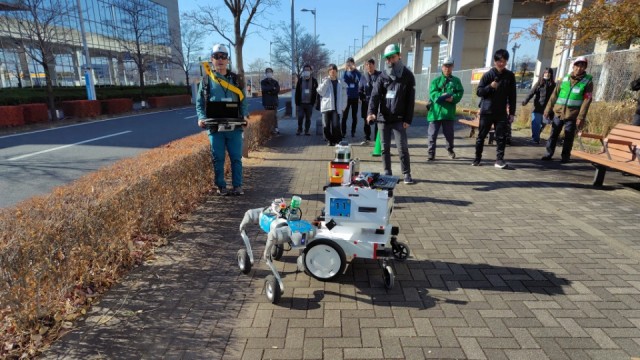}&
    \includegraphics[width=0.24\linewidth]{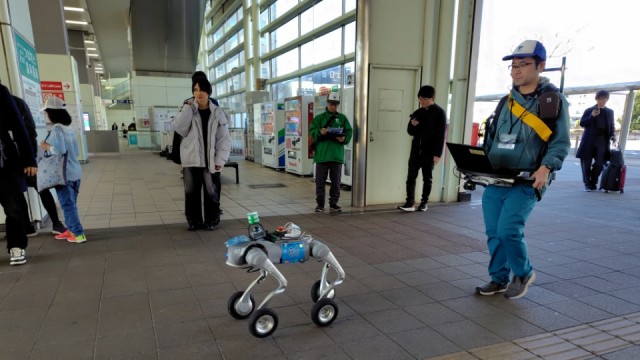}&
    \includegraphics[width=0.24\linewidth]{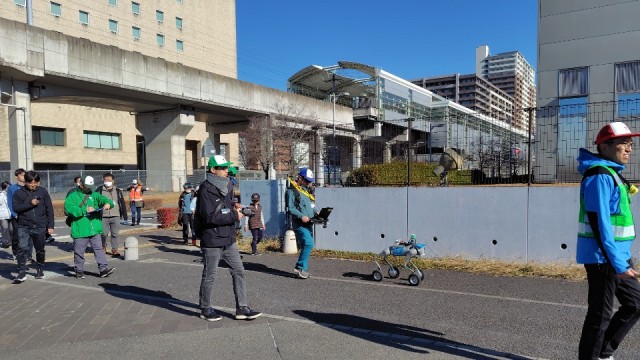}
    \\
    (i) & (j) & (k) & (l) \\[2mm]
    \includegraphics[width=0.24\linewidth]{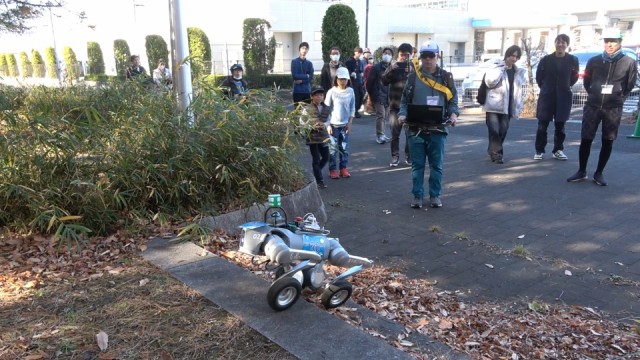}&
    \includegraphics[width=0.24\linewidth]{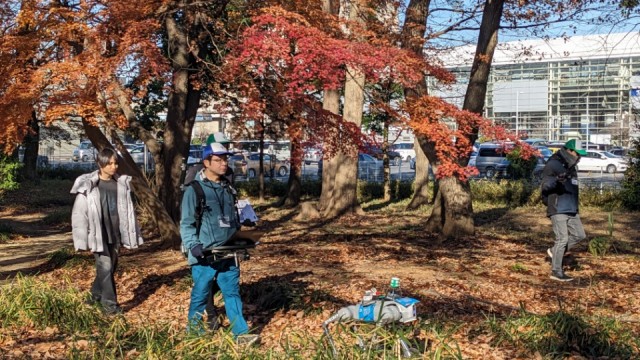}&
    \includegraphics[width=0.24\linewidth]{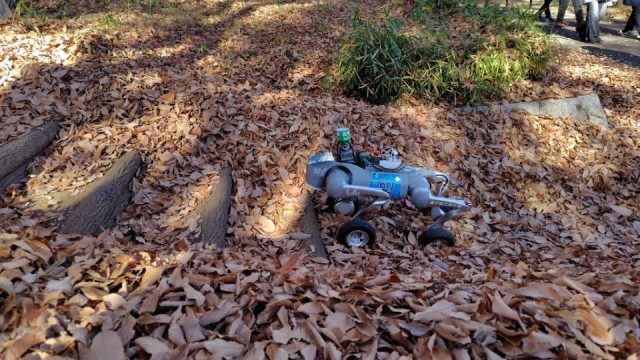}&
    \includegraphics[width=0.24\linewidth]{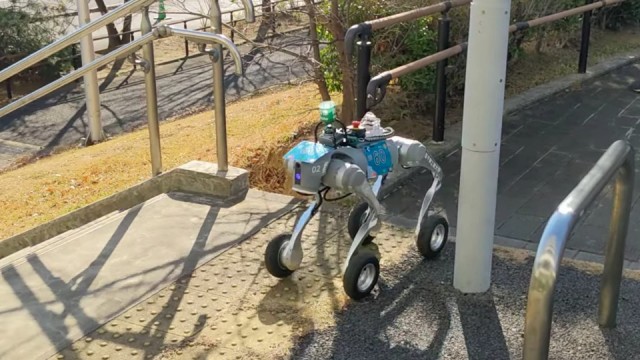}
    \\
    (m) & (n) & (o) & (p) \\[2mm]
    \includegraphics[width=0.24\linewidth]{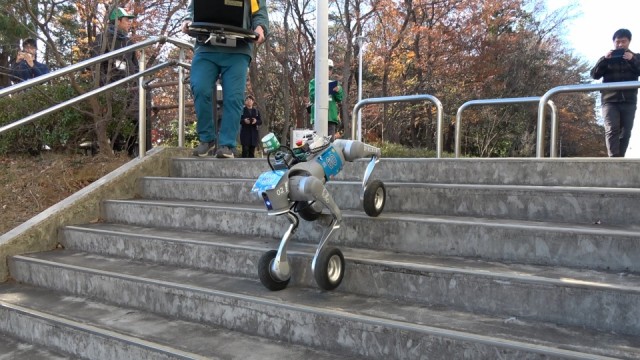}&
    \includegraphics[width=0.24\linewidth]{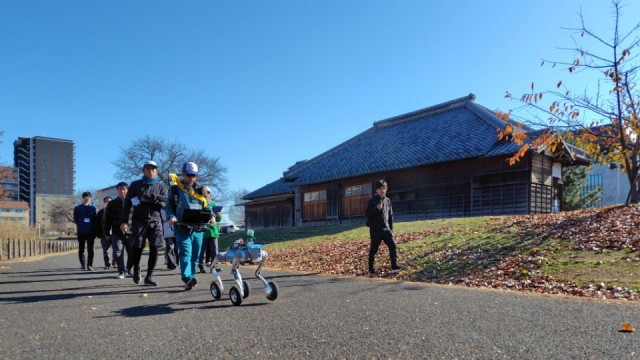}&
    \includegraphics[width=0.24\linewidth]{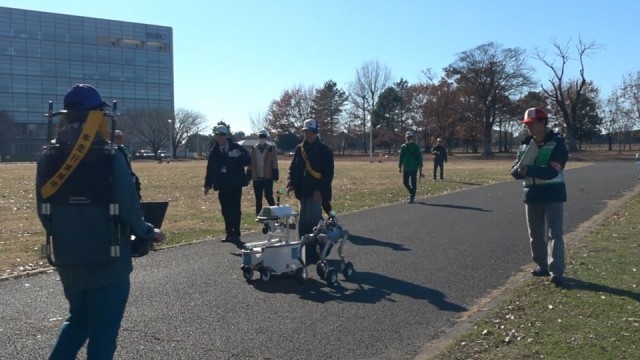}&
    \includegraphics[width=0.24\linewidth]{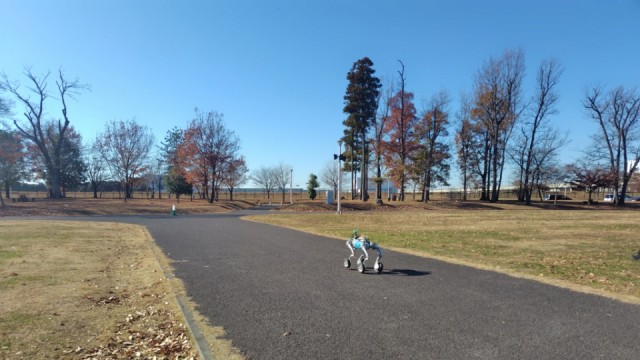}
    \\
    (q) & (r) & (s) & (t) \\[2mm]
    \includegraphics[width=0.24\linewidth]{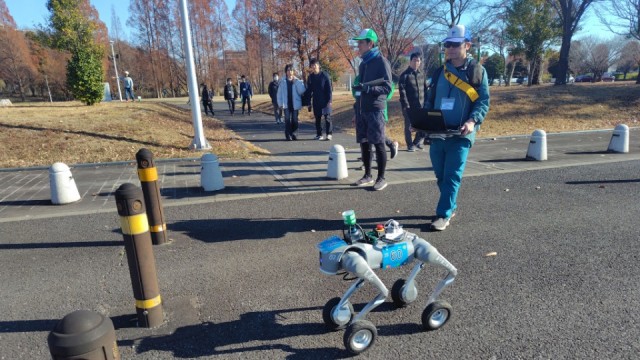}&
    \includegraphics[width=0.24\linewidth]{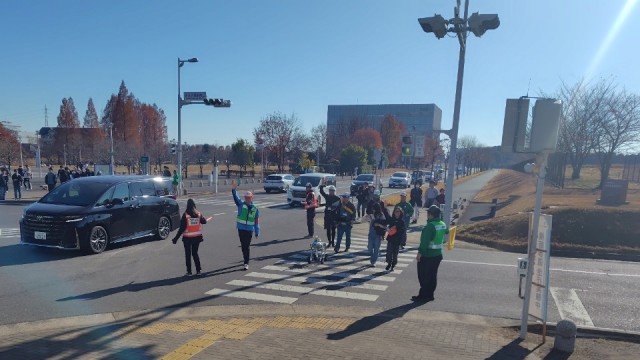}&
    \includegraphics[width=0.24\linewidth]{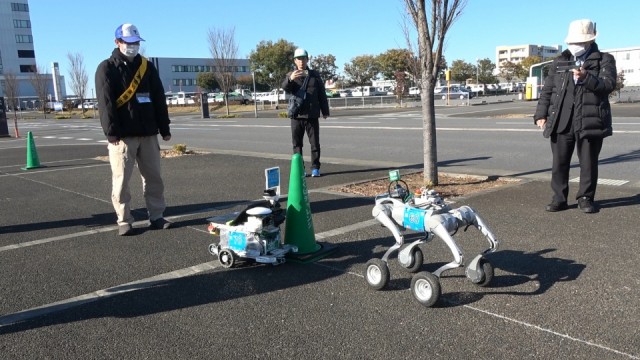}&
    \includegraphics[width=0.24\linewidth]{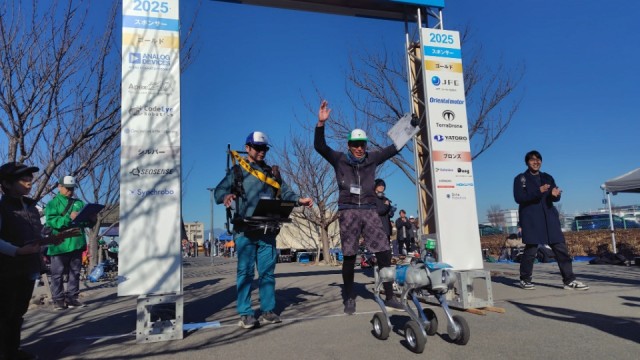}
    \\
    (u) & (v) & (w) & (x)
  \end{tabular}
  }
  \caption{
    Scenes from the final demonstration run on December 7.
    The robot started from the city hall (a), followed another robot on a path too narrow to overtake (b), and stopped at a crosswalk (f).
    At one point, it made contact with another robot (j), but both robots autonomously resumed navigation shortly after.
    It ascended stairs (m), including sections covered with fallen leaves (o), passed through a narrow gap between a streetlight and a handrail (p), and descended stairs (q).
    After avoiding obstacles in an open area (s), it reached the goal (x).
    A video of this run is available at \url{https://youtu.be/WQpMxAxA1eE}.
    }
  \label{fig:honsoukou_pics}
\end{figure*}

Fig.~\ref{fig:honsoukou_pics} shows scenes from the final demonstration run on December 7.
Unlike the earlier trial runs, a larger number of spectators gathered along the course, and many robots from other participating teams were navigating the route.
At one point, the robot came into physical contact with another robot (Fig.~\ref{fig:honsoukou_pics}(j)) but it continued navigating and completed the course without human intervention.
The battery level was 89\% at the start and 46\% at the end of the run.

\subsubsection{Motor Temperatures during the Runs}\label{sec:tc_temp}
\begin{figure*}[tb]
  \centering
  \includegraphics[width=\textwidth]{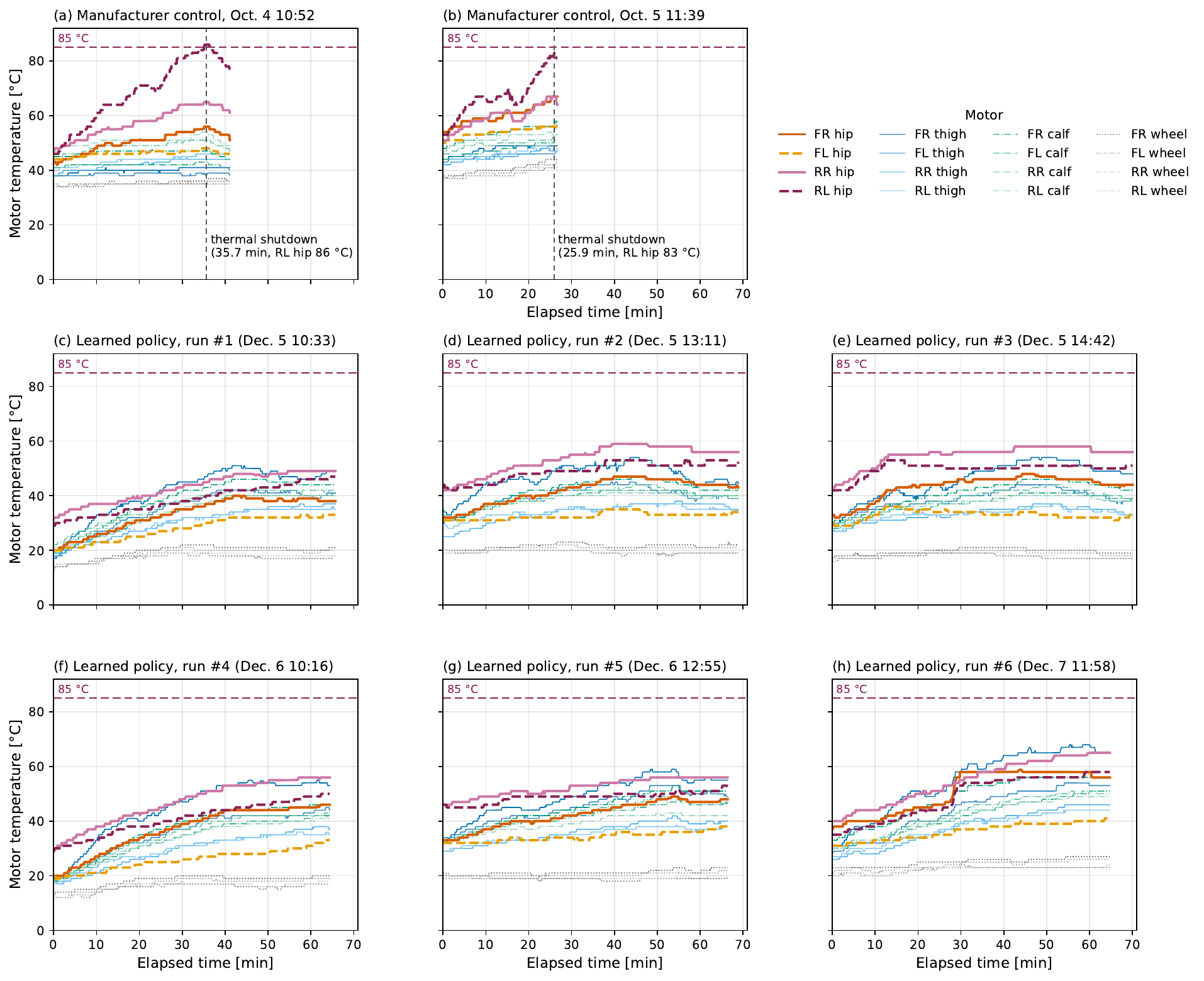}
  \caption{Temperatures of all 16 joint motors recorded by the onboard sensors during the runs on the Tsukuba Challenge 2025 course.
  (a), (b) Remote-controlled runs with the manufacturer-provided locomotion control on Oct.~4 and Oct.~5; the vertical dashed line marks the time at which the robot stopped due to the thermal protection.
  (c)--(h) Autonomous runs \#1--\#6 with the proposed system, which uses the flat-terrain policy except on the stair sections (Table~\ref{tab:TC_EXP_RES}).
  All panels share the same axes. The dashed horizontal line indicates \SI{85}{\degreeCelsius}, at which the thermal protection is triggered.
  Ambient temperature was \SIrange{25}{27}{\degreeCelsius} in October and \SIrange{7}{13}{\degreeCelsius} in December.}
  \label{fig:TC_TEMP_ALL}
\end{figure*}

Fig.~\ref{fig:TC_TEMP_ALL} shows the temperature of every joint motor during the runs on the Tsukuba Challenge course.
Plots (a) and (b) show the two runs in October with the manufacturer-provided control.
In both runs, the hip motor of the rear-left leg heated faster than the other motors. The robot came to a stop owing to thermal protection at 36 and 26 min from the start.
The manufacturer has not disclosed the threshold for thermal protection. However, in our experience, the robot stops when a motor reaches approximately \SI{85}{\degreeCelsius}.

Plots (c)--(h) show the six runs with the proposed system.
All six runs lasted more than one hour and covered approximately \SI{2.8}{\kilo\meter}.
The motor temperature at the start varied between \SI{30}{\degreeCelsius} and \SI{46}{\degreeCelsius} depending on the situation. Nevertheless, in every run, the temperature rise slowed down approximately 40 min after the start, and the maximum temperature of all motors remained within \SIrange{51}{68}{\degreeCelsius}.
This left a margin of more than \SI{17}{\degreeCelsius} from the \SI{85}{\degreeCelsius} threshold.
For a fair comparison, notably, the ambient temperature during the December runs was \SIrange{7}{13}{\degreeCelsius}. This was more favorable for cooling the motors than the \SIrange{25}{27}{\degreeCelsius} during the preliminary runs in October\footnote{Ambient temperatures are from the AMeDAS observations of the Japan Meteorological Agency at the Tsukuba (Tateno) station, \url{https://www.data.jma.go.jp/stats/etrn/} (accessed November 1, 2026).}.
In Fig.~\ref{fig:TC_TEMP_ALL}(h), the temperature rises sharply approximately 30 min after the start.
This was caused by physical contact with another autonomously running robot, during which the wheels of the two robots became entangled and the motors kept exerting torque until the robot became free.

\FloatBarrier

\subsection{Evaluation of Heat Reduction}
\label{sec:exp_heat}

We conducted an experiment to measure the effectiveness of our flat-terrain locomotion policy in suppressing motor heating in the same environment and under the same ambient temperature.
We compared the motor temperature rise during locomotion using the manufacturer-provided control (baseline) and the flat-terrain locomotion policy described in Section~\ref{subsec:locomotion_policy}.

The experiment was conducted in a flat outdoor environment at an ambient temperature of approximately \SI{28}{\degreeCelsius}, with a velocity command of \SI{1.0}{\meter\per\second} along a predetermined course.

\begin{figure}
  \centering
  \includegraphics[width=0.75\linewidth]{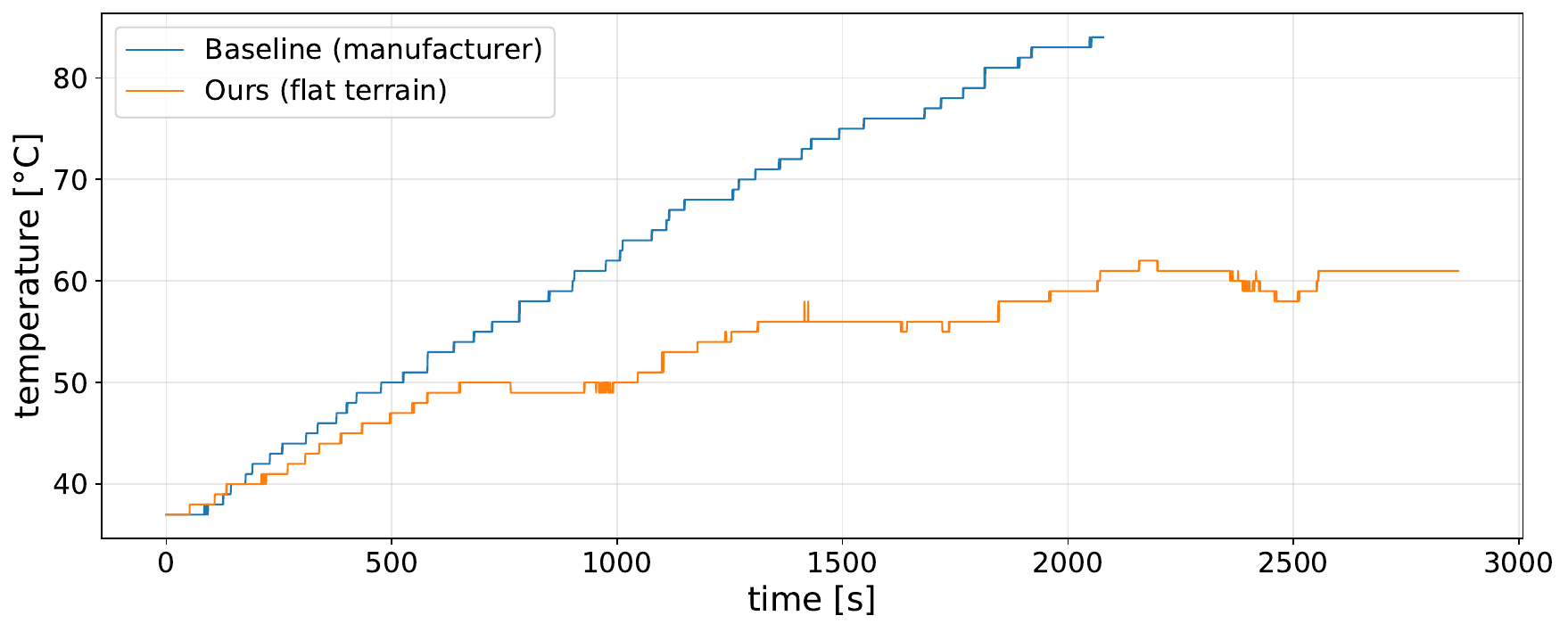}
  \caption{Maximum motor temperature during sustained locomotion at \SI{1.0}{\meter\per\second}. The time axes are aligned to the same starting temperature.
   The manufacturer-provided control (blue) reached 84\textdegree{}C at 34 min and stopped due to overheating, whereas our learned flat-terrain policy (orange) plateaued around 60\textdegree{}C.}
  \label{fig:TEMP_GRAPH}
\end{figure}

Fig.~\ref{fig:TEMP_GRAPH} shows the time history of the motor temperature during locomotion.
The graph plots the temperature of the hottest motor in the system.
With the manufacturer-provided control, the temperature rose continuously and reached \SI{84}{\degreeCelsius} at 34 min from the start, at which point the onboard controller of the robot triggered a shutdown owing to overheating.
The experiment was terminated at this point, resulting in the trace being truncated.

In contrast, with our flat-terrain locomotion policy, the temperature rise was gradual, and the temperature reached a plateau around \SI{60}{\degreeCelsius} approximately 40 min after the start.
The robot continued the locomotion for 51 min without triggering the thermal protection mechanism.

\section{Discussion}\label{sec:discussion}

\subsection{Analysis of Heat Reduction}
This subsection discusses the mechanism by which the flat-terrain locomotion policy suppresses heating.

\subsubsection{Analysis of Temperature and Torque Logs}
\begin{figure*}[tb]
  \centering
  \includegraphics[width=\textwidth]{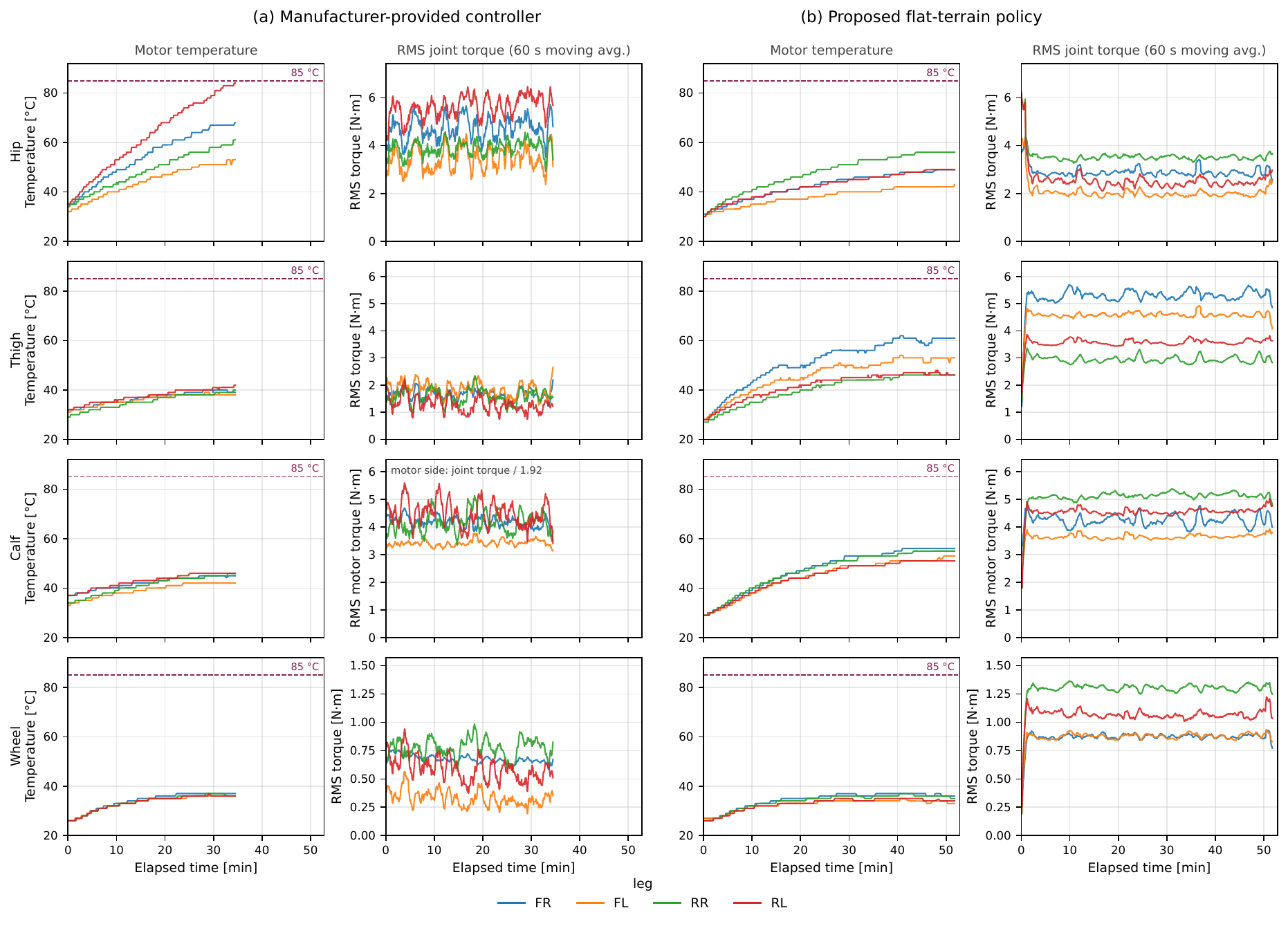}
  \caption{Motor temperature (left) and RMS joint torque (right, \SI{60}{\second} moving average of the torque estimates $\tau_{\mathrm{est}}$) of each joint group during the outdoor thermal test (Fig.~\ref{fig:TEMP_GRAPH}): (a) manufacturer-provided controller, (b) proposed flat-terrain policy.}
  \label{fig:CAMPUS_TEMP_TORQUE}
\end{figure*}

We analyzed the per-joint temperature rise in the experiment presented in Section~\ref{sec:exp_heat}.
Fig.~\ref{fig:CAMPUS_TEMP_TORQUE} shows the temperature of each joint motor (left columns) and root mean square (RMS) of the estimated joint torque smoothed by a \SI{60}{\second} moving average (right columns).
With the manufacturer-provided controller, the hip joints exhibited a pronounced temperature rise and reached the overheating threshold (\SI{85}{\degreeCelsius}).
In addition, the hip torque tended to increase with time during the run, which is considered to have caused the overheating.
In contrast, with the proposed flat-terrain policy, the hip, thigh, and calf joints approximately reached equilibrium within \SIrange{42}{62}{\degreeCelsius}, and a temperature rise concentrated in a specific joint was avoided.

\begin{table}[tb]
  \centering
  \caption{Time-averaged squared joint torque per motor, $\overline{\tau^2}$, of each joint group in the outdoor thermal test on campus (Fig.~\ref{fig:CAMPUS_TEMP_TORQUE}).}
  \label{tab:CAMPUS_TORQUE}
  \small
  \begin{tabular}{@{}llrrrrr@{}}
   & & \multicolumn{5}{c}{$\overline{\tau^2}$ per motor [\si{\newton\squared\meter\squared}]} \\
  Run & Control & Hip & Thigh & Calf & Wheel & Total \\ \hline
  (a) & Manufacturer & 20.5 & 2.7 & 16.7 & 0.4 & 40.3 \\
  (b) & Proposed & 7.8 & 17.7 & 19.7 & 1.1 & 46.3 \\
  \end{tabular}
  \normalsize
\end{table}

Table~\ref{tab:CAMPUS_TORQUE} summarizes the mean squared torque $\overline{\tau^2}$, which is the time average of the squared torque estimates while moving, for each joint group.
Notably, the torque values are estimates reported by the motor controller via the SDK, not direct measurements.
The calf joint is driven from the motor output shaft through a reduction linkage at the knee.
Based on the manufacturer's URDF model, we estimated the transmission ratio between the motor output shaft and the calf joint to be 1.917.
Therefore, the calf torque estimates reported by the SDK were divided by this ratio to convert them to the motor-side values before comparison with the other joint groups.
For the hip joints, $\overline{\tau^2}$ of the proposed policy is approximately 0.4 times that of the manufacturer-provided controller.
This is considered to be the main reason for the suppression of overheating.
In contrast, the torque at the other joints is larger with the proposed policy.
As a result, the total $\overline{\tau^2}$ over the four joint groups increases to approximately 1.15 times that of the manufacturer-provided controller.
These results indicate that the total load increases with the proposed policy.
However, the load concentrated at the hip joints is distributed to the other joints.
This distribution is considered to prevent the overheating of the hip motors and subsequent shutdown of the whole system.

\begin{table}[tb]
  \centering
  \caption{Time-averaged squared joint torque per motor, $\overline{\tau^2}$, of each joint group during the runs on the Tsukuba Challenge 2025 course, computed from the torque estimates ($\tau_{\mathrm{est}}$) while moving (odometry speed $>$ \SI{0.1}{\meter\per\second}).}
  \label{tab:TC_TORQUE}
  \small
  \begin{tabular}{@{}llrrrrr@{}}
   & & \multicolumn{5}{c}{$\overline{\tau^2}$ per motor [\si{\newton\squared\meter\squared}]} \\
  Run & Control & Hip & Thigh & Calf & Wheel & Total \\ \hline
  M1 & Manufacturer & 16.8 & 3.7 & 19.5 & 0.5 & 40.5 \\
  M2 & Manufacturer & 19.3 & 6.3 & 19.1 & 0.8 & 45.5 \\
  \hline
  \#1 & Proposed & 9.8 & 17.8 & 20.3 & 1.2 & 49.1 \\
  \#2 & Proposed & 11.9 & 19.4 & 20.3 & 1.2 & 52.8 \\
  \#3 & Proposed & 12.6 & 18.8 & 19.9 & 1.2 & 52.5 \\
  \#4 & Proposed & 11.7 & 19.1 & 20.0 & 1.2 & 52.0 \\
  \#5 & Proposed & 10.9 & 19.2 & 21.4 & 1.1 & 52.6 \\
  \#6 & Proposed & 12.2 & 19.6 & 20.1 & 1.3 & 53.2 \\
  \end{tabular}
  \normalsize
\end{table}

For reference, Table~\ref{tab:TC_TORQUE} shows the same statistics computed from the Tsukuba Challenge data in Section~\ref{sec:Tsukuba_Challenge} for the two runs with the manufacturer-provided controller and six runs with the proposed system.
As the robot stopped temporarily at traffic signals and crosswalks during the Tsukuba Challenge runs, only the periods in which the odometry speed exceeded \SI{0.1}{\meter\per\second} were included.
Here again, the total $\overline{\tau^2}$ over the four joint groups of the proposed policy was approximately 1.2 times that of the manufacturer-provided controller, confirming the same tendency.

\subsubsection{Comparison of Joint Behavior during Short Test Runs}
The internal specifications of the manufacturer-provided controller are not publicly available. Hence, the detailed mechanism responsible for the overheating could not be determined. To investigate this, we experimentally compared the behaviors of the manufacturer-provided controller and the proposed policy.
The same locomotion sequence (\SI{6}{m} forward, \SI{90}{\degree} right turn, \SI{3}{m} forward) was executed with both controllers, and the hip-joint angles and torque estimates reported by the SDK were recorded.
The results are shown in Fig.~\ref{fig:hip_angle_torque}.

\begin{figure}[tb]
  \centering
  \includegraphics[width=0.75\linewidth]{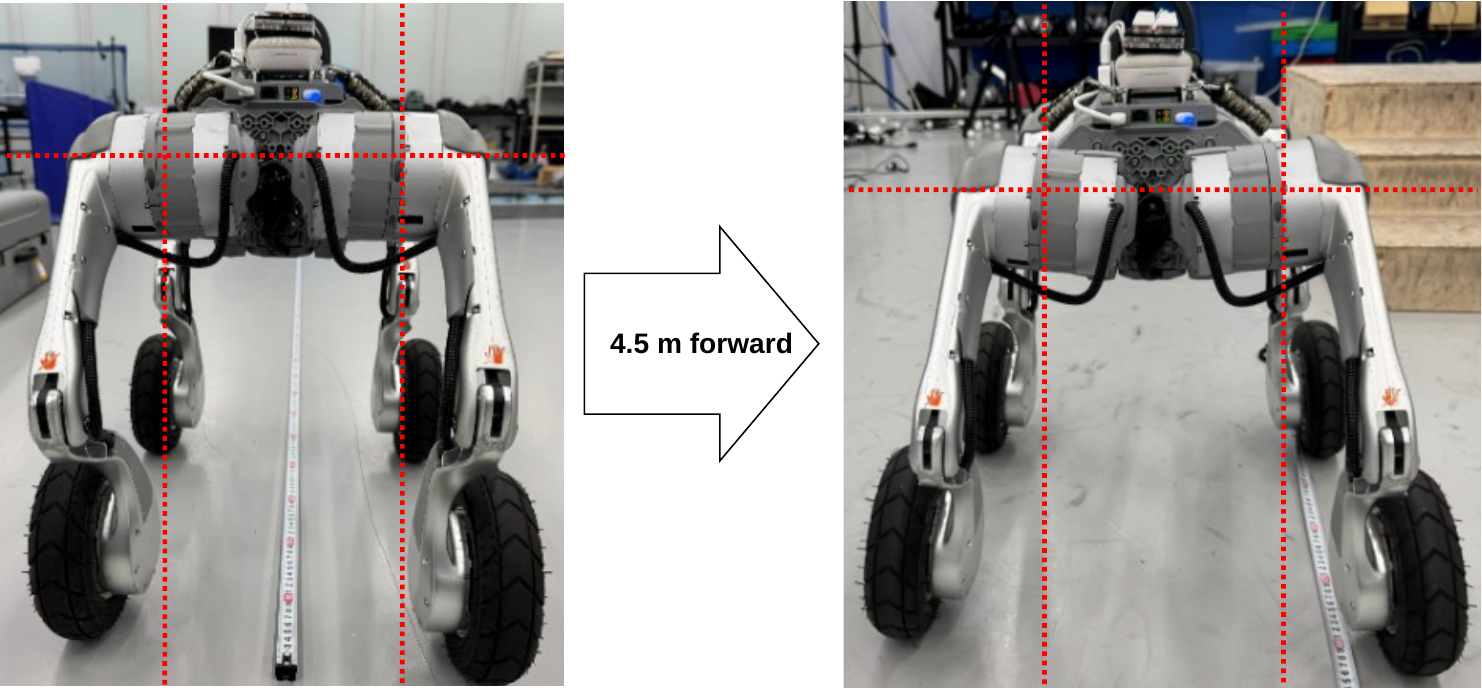}
  \caption{Leg posture under the manufacturer-provided controller at the start (left) and after \SI{4.5}{m} of straight travel (right).}
  \label{fig:leg_spread}
\end{figure}

With the manufacturer-provided controller (Fig.~\ref{fig:hip_angle_torque}(a)), the hip angle gradually shifts outward during straight-line travel, and a foot-lifting motion occurs during turning, which returns the angle to its original value.
Furthermore, the hip-joint torque increases over time during straight-line travel and decreases after the turn.
Fig.~\ref{fig:leg_spread} shows the leg posture after \SI{4.5}{m} of straight travel with the manufacturer-provided controller.
The photographs indicate that the hip angles have spread and tire contact points have moved outward from directly below the hip-joint axes.

In contrast, with the proposed flat-terrain locomotion policy (Fig.~\ref{fig:sim_gait}, Fig.~\ref{fig:hip_angle_torque}(b)), periodic foot-lifting motions occur even during straight-line travel.
As a result, the hip joints periodically return to their original angles instead of continuing to spread outward.
Similarly, the torque does not increase monotonically but varies periodically.

\subsubsection{Causes of Overheating and Effect of Foot Lifting}
\begin{figure*}[tb]
  \centering
  \tabcolsep=0pt
  \begin{tabular}{c}
    \includegraphics[width=0.75\linewidth]{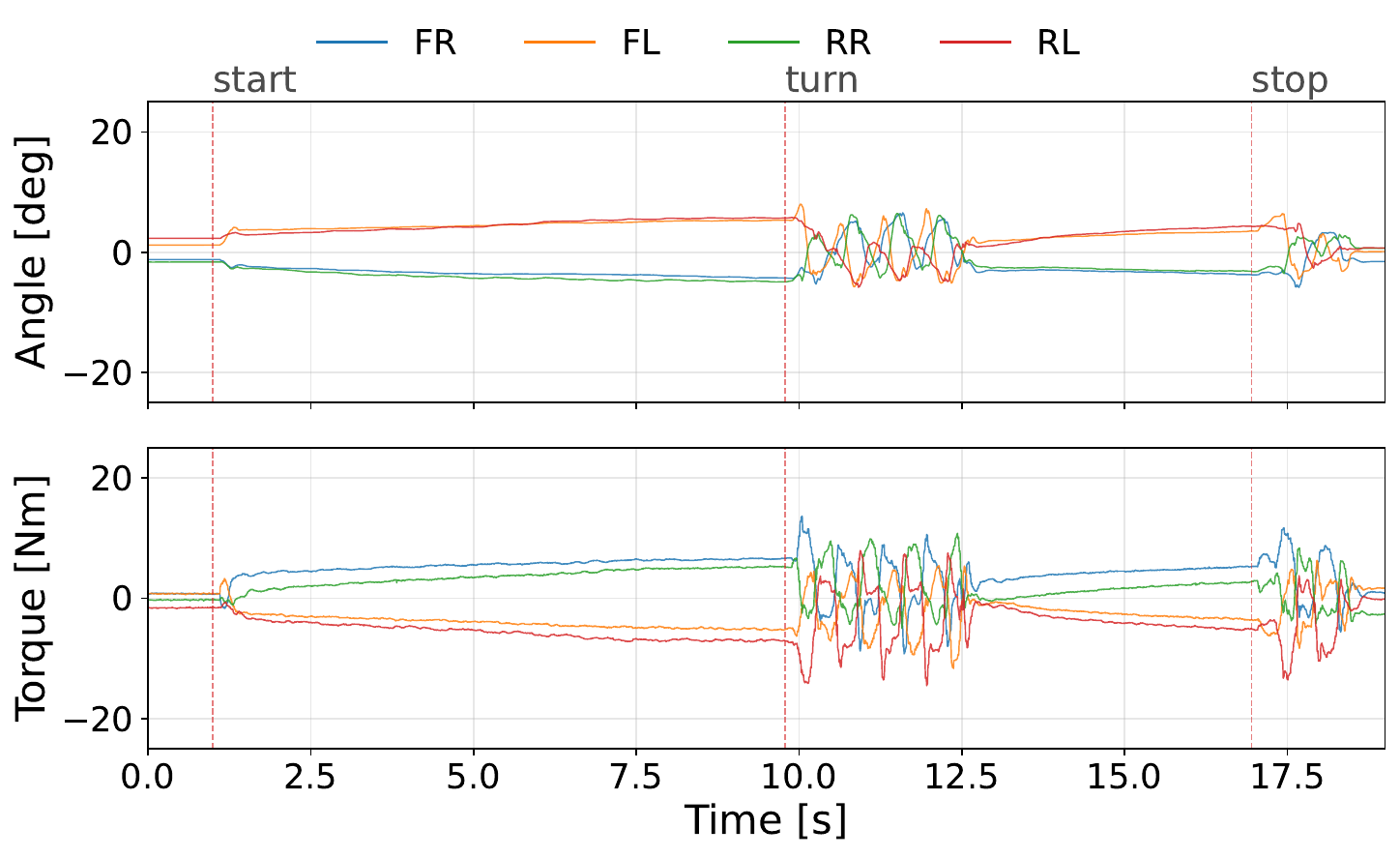} \\
    (a) Manufacturer gait \\ \\
    \includegraphics[width=0.75\linewidth]{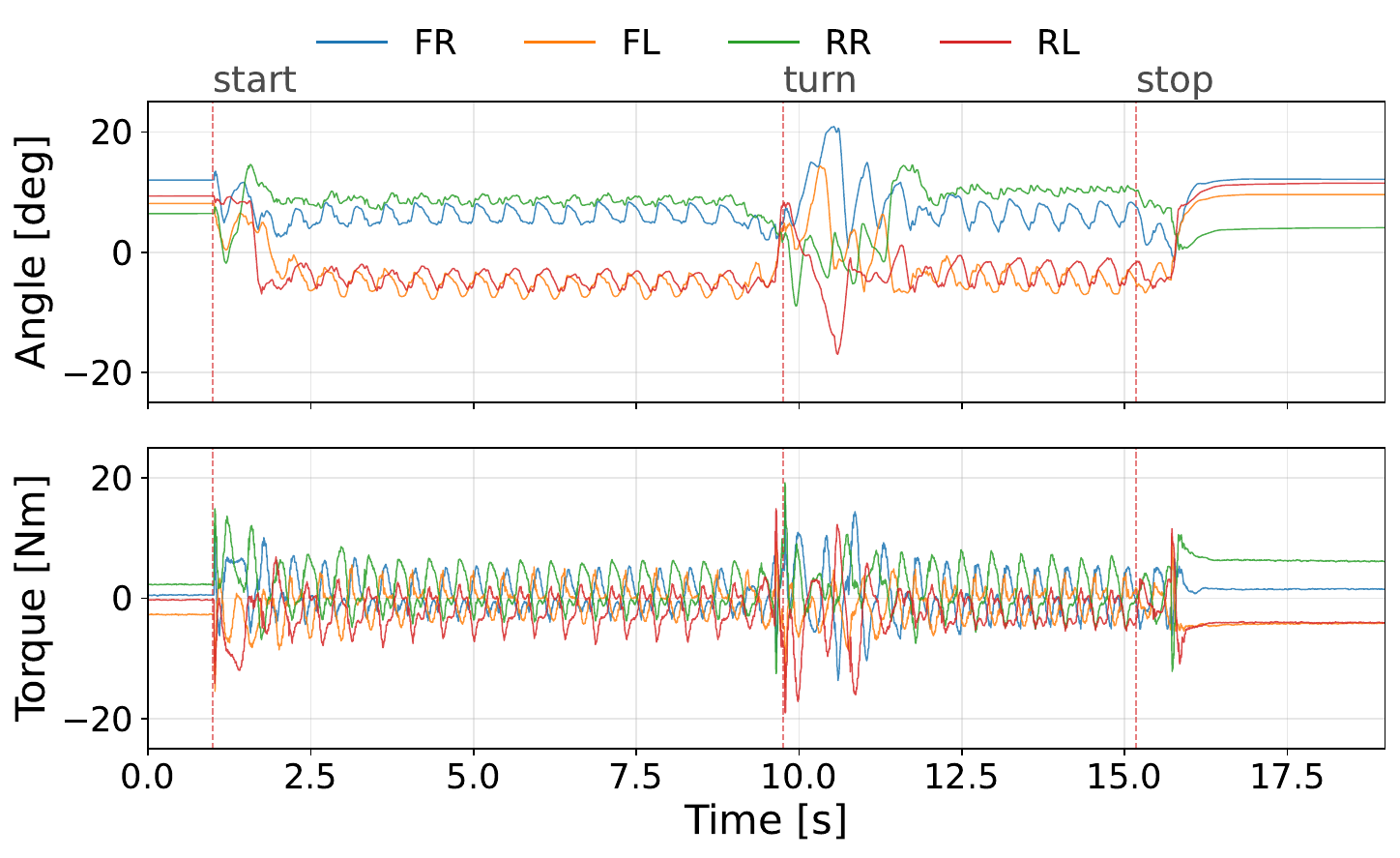} \\
    (b) Ours (flat terrain)
  \end{tabular}
  \caption{Hip joint angle and torque during a locomotion sequence consisting of \SI{6}{m} forward travel, a \SI{90}{\degree} right turn, and \SI{3}{m} forward travel.}
  \label{fig:hip_angle_torque}
\end{figure*}

From these observations, we infer the following three causes of overheating under the manufacturer-provided controller.
First, in the upright posture of Go2-W, the wheel contact point is located laterally outward from directly below the hip joint. Therefore, the body weight continuously exerts an outward moment on the legs.
Second, Go2-W has no degree of freedom that acts as steering for the wheels. While the wheels remain in contact with the ground, the only way to close the spread legs is for the hip-joint motors to exert a torque. However, as seen in Fig.~\ref{fig:hip_angle_torque}(a), the legs were not closed, presumably because of the large friction with the ground.
Third, the hip-joint motors are housed inside the body of the Go2-W. This is considered disadvantageous for cooling when compared with the other joint motors, which are exposed to the outside air.
The calf joints under the manufacturer-provided controller and thigh joints under the proposed policy produce torques of a magnitude close to those of the hip joints under the manufacturer-provided controller. Nevertheless, their temperature rise was small in both cases, which suggests a difference in the cooling conditions.

With the proposed flat-terrain policy, the periodic foot-lifting motion returns the wheel contact points to their original positions.
Accordingly, the torque demand is redistributed across the joints, which cool more easily, and overheating is considered to have been avoided.
This periodic foot-lifting motion is considered an effective strategy for distributing the load concentrated at a specific joint. It applies to legged-wheeled robots without steering degrees of freedom, such as the Go2-W.

In this study, temperature information was not directly used in policy training owing to the difficulty in incorporating it into the learning process.
The simulation of motor heating requires not only estimating the heat generation from the electrical power and torque but also modeling the internal heat conduction and dissipation to the ambient environment, which is beyond the scope of standard dynamics simulation.
Moreover, the heating process occurs over tens of minutes, rendering it difficult to reproduce within a training episode.
Therefore, we adopted the approach of indirectly encouraging the foot-lifting behavior through adjustments to the terrain parameters and reward functions.
A recent study has proposed directly modeling the heating through simulation and incorporating the temperature into the policy training~\cite{wan2026thermal_rl}, which is a direction worth exploring.

\subsection{Limitations and Scope of Future Research}
In this study, we demonstrated that the developed system is capable of long-distance navigation in real-world environments.
However, the system has certain limitations, as discussed below. In addition, we present the directions for future research.

First, a trade-off exists between thermal suppression and rough-terrain traversal capability.
As described in Section~\ref{sec:locomotion}, we have not yet achieved simultaneous thermal suppression and high traversal capability within a single policy;
instead, a flat-terrain locomotion policy with reduced heating and rough-terrain locomotion policy with high traversal capability but limited operational duration owing to heating are alternately used.
Moreover, this policy switching currently relies on manual configuration; the ability of the robot to autonomously select an appropriate policy based on the terrain has not been implemented.
Future research will explore methods to learn a single policy that achieves both thermal suppression and traversal capability, as well as automate terrain-dependent policy selection.

Second, the navigation system relies on classical perception and planning methods and can be further improved.
The current system handles obstacles through deceleration, stopping, and path replanning, but does not perform avoidance based on identification or prediction of moving obstacles.
In experiment \#3 in Table~\ref{tab:TC_EXP_RES}, the accompanying operator had to press the emergency stop button when an oncoming vehicle appeared from a blind spot.
Incorporating approaches such as occlusion-aware motion planning~\cite{kiybib:zhang_rss2021_occlusion_aware_driving,kiybib:firoozi_tcst2025_oa_mpc} is another direction for future research.
In addition, the system relies on a pre-built 3D point cloud map and predefined route, and cannot operate in environments without prior preparation.
Introducing semantic traversability estimation and online path planning~\cite{mattamala2025wvn,kim2024semantic_traversability} to enable flexible navigation that considers the traversability of sidewalks, roadways, vegetation, and steps is a future challenge.

Finally, this study has not demonstrated that the performance of the legged-wheeled configuration surpasses that of wheeled or legged robots in practice.
Comparative evaluations of the locomotion speed and energy efficiency across various environments with robots of other configurations would help clarify the advantages and disadvantages of the legged-wheeled configuration and is an important direction for future research.

\section{Conclusion}\label{sec:conclusion}
In this study, we developed a locomotion control and autonomous navigation system based on DRL for the commercially available legged-wheeled robot Go2-W.
We extended the locomotion control originally designed for quadruped robots to the legged-wheeled robot, obtained a locomotion policy that addresses the heat concentration problem during wheeled locomotion,
and incorporated a path-following controller adapted to the locomotion characteristics of legged-wheeled mobility.
In the Tsukuba Challenge 2025, the robot completed an approximately \SI{2.8}{\kilo\meter} course that included sidewalks, a park, stairs, and unpaved terrain, demonstrating the feasibility of long-distance outdoor autonomous navigation with a legged-wheeled robot.

Although this study targeted a specific robot, Go2-W, we expect that the approach for training locomotion policies that suppress heating, analysis of the heat concentration mechanism, and navigation system design will be informative for researchers working on legged-wheeled robots with similar configurations.
Future research includes investigating learning methods that achieve both thermal management and terrain traversal capability and addressing dynamic obstacles.

\section*{Appendix A: Details of Policy Training}\label{sec:appendix}

This appendix provides supplementary details on the training of the locomotion policies.

\subsection*{Training Parameters}

Table~\ref{tab:training_params_combined} lists the main parameters used for training each policy.
The control period of each policy is given by the product of the simulation time step $\Delta t$ and decimation factor, yielding $0.02$\,s (\SI{50}{\hertz}) for all policies.

\begin{table}[tb]
    \centering
    \caption{Training parameters used for training each policy.}
    \label{tab:training_params_combined}
    \small
    \begin{tabular}{@{}lccc@{}}
    \toprule
    Parameter & Stance & Rough terrain & Flat terrain \\
    \midrule
    Number of iterations       & 4{,}000  & 40{,}000 & 20{,}000     \\
    Episode length [s]         & 10       & 40            & 40           \\
    $x$ velocity command range [m/s]     & 0        & $\pm 0.6$     & $\pm 1.0$    \\
    $y$ velocity command range [m/s]     & 0        & $\pm 0.3$     & $\pm 0.5$    \\
    $\omega$ command range [rad/s]     & 0        & $\pm 1.0$     & $\pm 0.5$    \\
    Joint PD gain $K_p$        & 50       & 50            & 40           \\
    Joint PD gain $K_d$        & 1.0      & 1.0           & 1.0          \\
    Action scale               & 0.25     & 0.25          & 0.25         \\
    Wheel speed scale          & 4.0      & 4.0           & 6.0          \\
    Dynamic friction           & 1.0      & 0.7           & 1.0          \\
    Simulation $\Delta t$ [s]  & 0.005    & 0.005         & 0.004        \\
    Simulation substeps        & 1        & 1             & 2            \\
    Control decimation         & 4        & 4             & 5            \\
    \botrule
    \end{tabular}
\end{table}

Table~\ref{tab:domain_rand} lists the domain randomization~\cite{Tobin2017DomainRF} parameters.
The added base mass was varied from \SI{-1}{\kilogram} to \SI{5}{\kilogram} to ensure robustness to the weight of additional equipment mounted for navigation.

\begin{table}[tb]
    \centering
    \caption{Domain randomization parameters (common to all policies).}
    \label{tab:domain_rand}
    \small
    \begin{tabular}{@{}lc@{}}
    \toprule
    Parameter & Range \\
    \midrule
    Ground friction coefficient    & $[0.05,\; 1.5]$ \\
    Added base mass [kg]           & $[-1.0,\; 5.0]$ \\
    Action latency [steps]         & $[3,\; 8]$ \\
    \cmidrule{1-2}
    \multicolumn{2}{@{}l}{\textit{Observation noise scales}} \\
    \quad Joint positions [rad]    & 0.01 \\
    \quad Joint velocities [rad/s] & 1.5 \\
    \quad Angular velocity [rad/s] & 0.2 \\
    \quad Gravity direction        & 0.05 \\
    \botrule
    \end{tabular}
\end{table}

\subsection*{Custom Reward Functions}

Legged Gym~\cite{RudinHR021} provides a set of reward functions for training locomotion policies for legged robots.
We have previously introduced custom reward functions to improve the policy performance~\cite{kiybib:irie_ar_2025_qrc}.
This subsection describes the reward functions newly introduced in this study, which are marked with $^{**}$ in Table~\ref{tab:reward_functions}.

\subsubsection*{leg\_effort\_std}

This reward function penalizes imbalanced torque distribution across the four legs, encouraging load equalization.

First, an EMA filter is applied to each joint torque $\tau_j$ to obtain the smoothed torque $\bar{\tau}_j$.
\begin{equation}
  \bar{\tau}_{j,t} = w \, \bar{\tau}_{j,t-1} + (1 - w) \, \tau_{j,t},
\end{equation}
where $w$ is the EMA weight, set to $w=0.975$ for the flat-terrain policy and $w=0.7$ for the rough-terrain policy.
Note that this EMA is used for torque smoothing and is distinct from the EMA filters applied to the observation input of the flat-terrain policy described in Section~\ref{subsec:locomotion_policy}.

The smoothed torques $\bar{\tau}_j$ are then rearranged into a $4 \times K$ layout (four legs $\times$ $K$ joints per leg), and the deviation across the four legs is computed for each joint position $k$.
\begin{equation}
  R_\mathrm{leg\_effort\_std} = \sum_{k=1}^{K} \sqrt{\sum_{i=1}^{4} \left(\bar{\tau}_{i,k} - \frac{1}{4}\sum_{l=1}^{4}\bar{\tau}_{l,k}\right)^2},
\end{equation}
where $\bar{\tau}_{i,k}$ is the EMA-smoothed torque at the joint position $k$ of leg $i$.
A larger value indicates that the load is concentrated on specific legs.

The support\_phase\_ratio\_std from~\cite{kiybib:irie_ar_2025_qrc} evaluates the standard deviation of the stance-phase ratios based on the contact states, whereas leg\_effort\_std directly evaluates the load equalization based on the torques.

\subsubsection*{action\_curvature}

This reward function penalizes the curvature of the action sequence, promoting smooth periodic motions.
The discrete curvature is computed from the first- and second-order differences of the actions.
\begin{align}
  \dot{a}_{j,t} &= a_{j,t} - a_{j,t-1} \\
  \ddot{a}_{j,t} &= \dot{a}_{j,t} - \dot{a}_{j,t-1} .
\end{align}
The reward function is defined based on the following curvature formula:
\begin{equation}
  R_\mathrm{action\_curvature} = \sum_{j=1}^{N_\mathrm{dof}} \frac{|\ddot{a}_{j,t}|}{(1 + \dot{a}_{j,t}^2)^{3/2}}.
\end{equation}
The denominator $(1 + \dot{a}_{j,t}^2)^{3/2}$ reduces the sensitivity when the rate of change is large.
This means that large motions at a constant rate incur a small penalty, and only abrupt changes in the direction are strongly penalized.
In contrast to the action\_rate term included in Legged Gym~\cite{RudinHR021}, which uses the squared first-order difference, action\_curvature uses the second-order difference to more selectively suppress oscillatory behavior.

\subsection*{Ablation of Factors Inducing Foot Lifting}

We conducted an ablation study to examine which design elements contribute to the foot-lifting gait. The elements examined were the reward terms introduced in this study, leg\_effort\_std and action\_curvature (Table~\ref{tab:reward_functions}), and change in the representation of past observations from the last action to EMA-filtered observations.
Each item that had been changed for the flat-terrain policy was reverted to its previous setting (the same as applied for the rough-terrain policy), and the policy was trained under each condition.
For each condition, three policies were trained with different random seeds.
To keep the computation time manageable, the number of training iterations was limited to 10{,}000 here; therefore, the resulting policies differ slightly in performance from the deployed ones.
For a consistent comparison, the proposed flat-terrain and rough-terrain policies were also retrained under the same conditions.
The resulting policies were then run on a flat ground in the simulator at a forward velocity of \SI{0.4}{\meter\per\second} with three robots (one per seed) for \SI{60}{\second}. The duty factor and foot clearance were measured during these runs.

\begin{figure}[tb]
  \centering
  \includegraphics[width=0.75\linewidth]{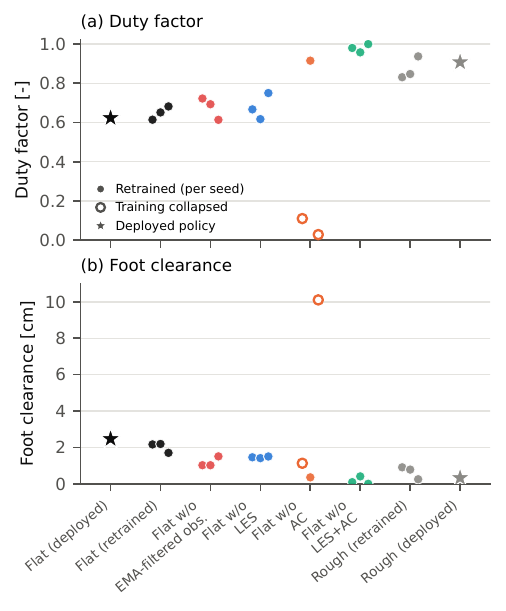}
  \caption{(a) Duty factor and (b) foot clearance in the reward-term ablation.
  The duty factor is the fraction of the run time during which each wheel is in contact with the ground, averaged over the four wheels.
  The foot clearance is the peak wheel-axle height relative to its stance height.
  Circles show each seed of the retrained policies and stars show the deployed policies.
  LES: leg\_effort\_std; AC: action\_curvature; EMA-filtered obs.: EMA-filtered past observations.
  Open circles mark seeds whose training collapsed (Fig.~\ref{fig:ablation_curves}); these policies oscillated and did not move forward properly.}
  \label{fig:fig_ablation_df}
\end{figure}

\begin{figure*}[tb]
  \centering
  \includegraphics[width=\textwidth]{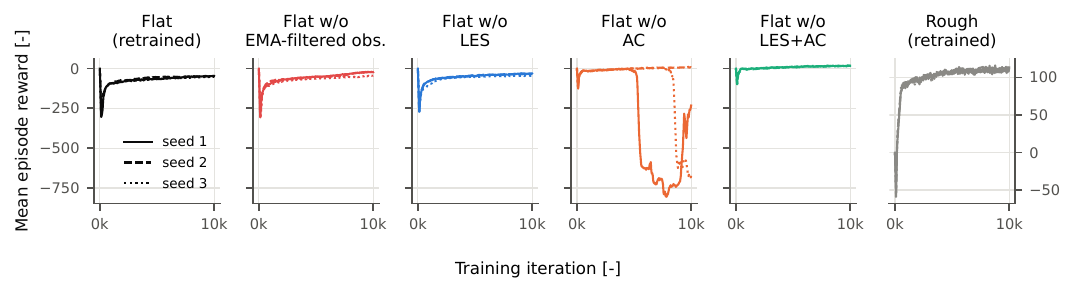}
  \caption{Learning curves of the reward-term ablation (three seeds per condition; 50-iteration moving average of the mean episode reward; abbreviations as in Fig.~\ref{fig:fig_ablation_df}). The Rough (retrained) panel has its own scale (right axis).}
  \label{fig:ablation_curves}
\end{figure*}

Fig.~\ref{fig:fig_ablation_df}(a) shows the duty factor and Fig.~\ref{fig:fig_ablation_df}(b) shows the foot clearance during the swing phase for each condition.
Reverting leg\_effort\_std or the past-observation representation alone did not eliminate the foot-lifting gait.
When only action\_curvature was reverted, one seed converged to a wheel-based gait and the other two collapsed during training (Fig.~\ref{fig:ablation_curves}).
When both leg\_effort\_std and action\_curvature were reverted, all three seeds converged to a wheel-based gait.
These results suggest that applying leg\_effort\_std and action\_curvature together yields the foot-lifting gait and improves the stability of training.
In addition, the EMA-filtered past observations were found to contribute to a larger foot clearance.

\section*{Appendix B: Details of Path-Following Control}\label{sec:appendix_b}

\subsubsection*{Learning the Movement Model}

To train the path-following policy, a simulator that reflects the movement of the robot in response to velocity commands was constructed from the data.
The motion data of the robot for various velocity commands were collected in advance and an estimator was trained through supervised learning, in a manner similar to that in Section~\ref{sec:navigation}.
The main difference from the odometry estimation model was that the output included the yaw angle change $\Delta\theta$, thereby predicting a 3-DoF relative movement $\Delta \mathbf{x}_{t+1} = \mathbf{x}_{t+1} \ominus \mathbf{x}_{t}$.
In addition, the relative movements from the past two steps were used as additional inputs along with the velocity commands.
Here, $\ominus$ denotes the operator that computes the difference between two poses, including rotation~\cite{kiybib:Lu_graphSLAM}.
Specifically, the movement model $f_\phi$ takes as input the current velocity command $\mathbf{u}_t$ and state $\mathbf{s}_t = [\Delta \mathbf{x}_{t}^{\rm T}, \Delta \mathbf{x}_{t-1}^{\rm T}, \mathbf{u}_{t-1}^{\rm T}, \mathbf{u}_{t-2}^{\rm T}]^{\rm T}$, and outputs the next relative movement, as follows:
\[
  \hat{\Delta \mathbf{x}}_{t+1} = f_\phi(\mathbf{s}_t, \mathbf{u}_t).
\]
The network architecture and training procedure are the same as those of the odometry estimation model.
By using this model as a lightweight simulator, the path-following policy can be trained in a short time without running a full 3D simulator.

\subsubsection*{Learning the Path-Following Policy}

The path-following policy was trained as a function that outputs velocity commands $(v_x, v_y, \omega)$ to the lower-level locomotion controller, given the current robot state and a local target point.
The local target point is a lookahead point set \SI{1.0}{\meter} ahead of the current position along the path.
The position and orientation of the lookahead point, given in the map coordinate frame, are transformed into the local coordinate frame of the robot and denoted as $\mathbf{g}_t$.
The path-following policy $\pi_\theta$ takes as input the same history state $\mathbf{s}_t$ as that of the movement model and target pose $\mathbf{g}_t$, and outputs the velocity command $\mathbf{u}_t = (v_{x,t}, v_{y,t}, \omega_t)^{\rm T}$.
The policy function is an MLP with two hidden layers of 256 units each, trained using Soft Actor Critic~\cite{haarnoja2018sac}.
\[
  \mathbf{u}_t = \pi_\theta(\mathbf{s}_t, \mathbf{g}_t) .
\]
During operation, the lookahead point is updated as the robot moves.
This can be viewed as a form of receding horizon control in which online optimization is replaced by a trained policy, which is expected to follow the path while preventing error accumulation caused by model inaccuracies.

The policy parameters $\theta$ are trained to maximize the cumulative reward obtained on the movement model.
\[
  \theta^* =
  \mathop{\rm arg\,max}_{\theta}
  \mathbb{E}_{\pi_\theta, f_{\phi^*}}
  \left[
    \sum_{t=0}^{T-1} R_t
  \right].
\]
The reward function is defined as follows:
\[
  \begin{array}{rl}
  R_t =
  & -\left\| \mathbf{g}_t \right\| \\
  & -\alpha_1 \max(\mathbf{u}_t^{\rm T}\mathbf{u}_t - 1,\; 0) \\
  & -\alpha_2 \left\| \mathbf{u}_t - \mathbf{u}_{t-1} \right\| \\
  & -\alpha_3 \left| g_{y,t} \right| \\
  & -\alpha_4 \left| g_{\theta,t} \right| \\
  & -1
  \end{array}.
\]
Here, $g_{y,t}$ and $g_{\theta,t}$ denote the lateral deviation and angular deviation from the target point, respectively, and $\alpha_1,\ldots,\alpha_4$ are the weights for each term.
The lateral and angular deviations are provided as separate terms to allow adjustment of the balance between turning and lateral stepping.
The constant term $-1$ is a per-step penalty that encourages the policy to minimize the number of steps required to reach the target point.
In the experiments, the weights were set to $\alpha_1 = 1.0$, $\alpha_2 = 3.0$, $\alpha_3 = 0.5$, and $\alpha_4 = 0.5$.

\subsubsection*{Comparison with Pure Pursuit}

\begin{figure}[tb]
  \centering
  \includegraphics[width=0.75\linewidth]{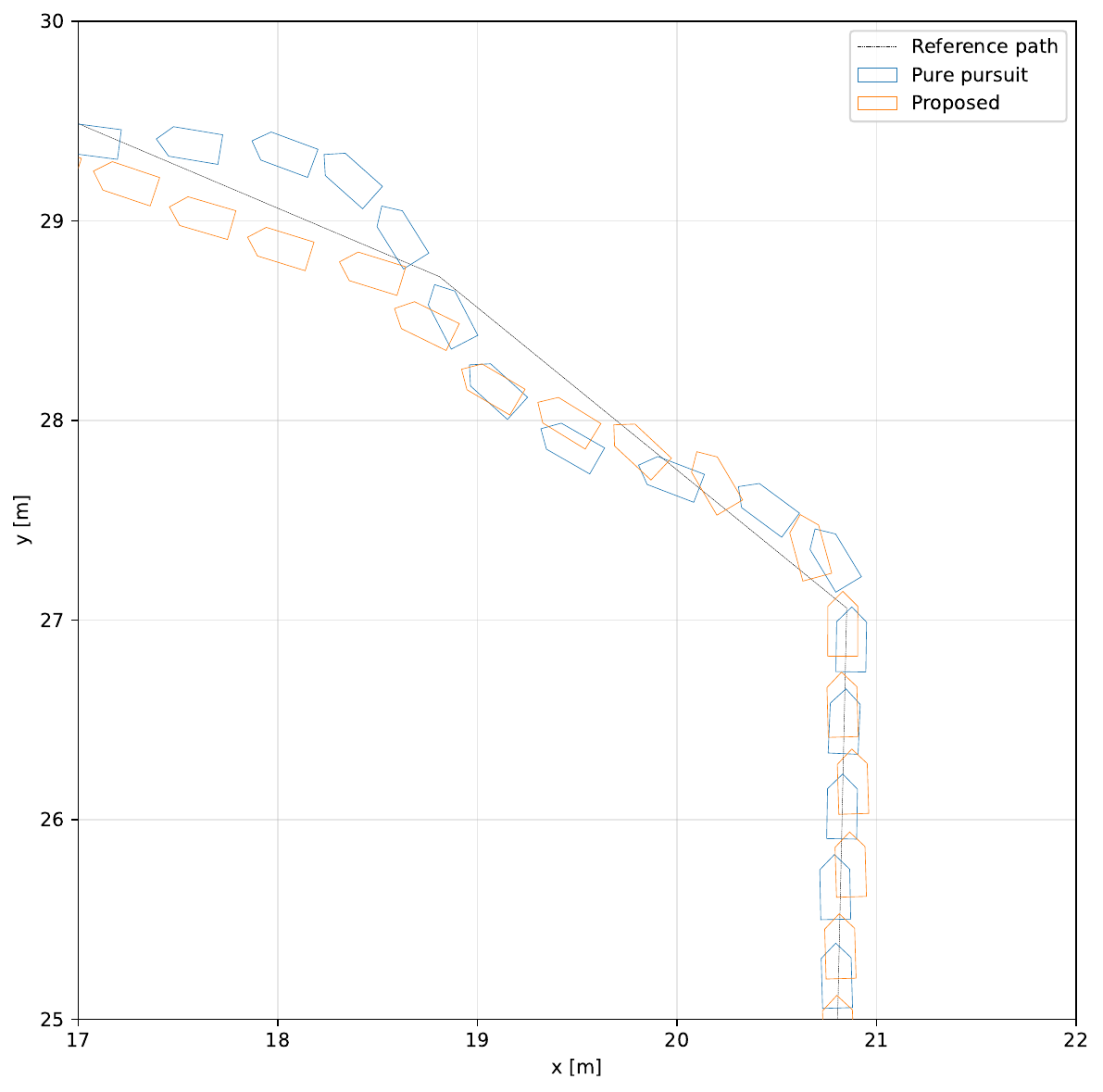}
  \caption{Comparison of path tracking between the proposed method (orange) and pure pursuit (blue). The reference path is shown as a black dashed line and the pentagons represent the robot footprint. Neither method was tuned on the real robot.}
  \label{fig:path_following_comparison}
\end{figure}

Fig.~\ref{fig:path_following_comparison} compares the tracking trajectories of the proposed method and pure pursuit on the real robot.
Pure pursuit exhibits oscillatory behavior, whereas the proposed method takes advantage of omnidirectional movement to achieve smoother path tracking.

\backmatter

\bmhead{Acknowledgements}
The authors used Claude (Anthropic, Inc.) as an aid in translating the Japanese draft into English and in editing the manuscript.
All text was reviewed and revised by the authors.
The manuscript was further edited by Editage (http://www.editage.jp), an English language editing service.

\section*{Declarations}
\begin{itemize}
\item \textbf{Funding} Not applicable.
\item \textbf{Conflict of interest} The authors declare no competing interests.
\item \textbf{Data availability} The 3D point cloud map used in this study is publicly available at the Tsukuba Challenge dataset repository. Experimental data for temperature and navigation are available from the corresponding author upon reasonable request. A full video of the navigation run at the Tsukuba Challenge is available at \url{https://youtu.be/WQpMxAxA1eE}.
\item \textbf{Code availability} A reference implementation of the localization module is available at \url{https://github.com/kiyoshiiriemon/simple_fastlio_localization}.
\end{itemize}

\bibliography{references}

\end{document}